\pdfoutput=1
\documentclass{article}
\pdftrailerid{}
\usepackage{iclr2027_conference,times}
\iclrfinalcopy
\usepackage{amsmath,amssymb,graphicx,booktabs,url,xcolor,capt-of,wrapfig}
\usepackage{array,colortbl,needspace}

\definecolor{bandblue}{RGB}{214,227,242}
\definecolor{keyblue}{RGB}{235,242,251}
\definecolor{accgreen}{RGB}{226,240,227}
\definecolor{ourblue}{rgb}{0.368,0.507,0.71}
\newcolumntype{K}{>{\columncolor{keyblue}}c}
\newcolumntype{J}{>{\columncolor{keyblue}}l}
\newcommand{\band}{\rowcolor{bandblue}}
\newcommand{\keyrow}{\rowcolor{accgreen}}
\usepackage[utf8]{inputenc}
\usepackage{hyperref}
\hypersetup{
    colorlinks,
    linkcolor=ourblue,
    citecolor=ourblue,
    urlcolor=ourblue
}

\title{%
  Refusal Localizes, the Damage Relocates: Safety Layers Under Few-Sample Fine-Tuning%
  \vspace{14pt}%
}

\author{%
  \vspace*{4pt}
  \makebox[5.28in][c]{%
  \begin{tabular}{@{}c@{\hspace{0.02in}}c@{\hspace{0.02in}}c@{}}
    \parbox[c]{1.76in}{\centering Jungseob Lee$^{1,*}$\\[1pt] \fontsize{8pt}{7.5pt}\texttt{omanma1928@korea.ac.kr}} &\vspace*{4pt}
    \parbox[c]{1.76in}{\centering Dongyub Jude Lee$^{2,*}$\\[1pt] \fontsize{8pt}{7.5pt}\texttt{jude.lee@zoom.us}} &\vspace*{4pt}
    \parbox[c]{1.76in}{\centering Sugyeong Eo$^{3,\dagger}$\\[1pt] \fontsize{8pt}{7.5pt}\texttt{s.eo@yonsei.ac.kr}}\\[4pt]\vspace*{4pt}
    \parbox[c]{1.76in}{\centering Seongtae Hong$^{1}$\\[1pt] \fontsize{8pt}{7.5pt}\texttt{ghdchlwls123@korea.ac.kr}} &
    \parbox[c]{1.76in}{\centering Seungyoon Lee$^{1}$\\[1pt] \fontsize{8pt}{7.5pt}\texttt{dltmddbs100@korea.ac.kr}} &
    \parbox[c]{1.76in}{\centering Heuiseok Lim$^{1,\dagger}$\\[1pt] \fontsize{8pt}{7.5pt}\texttt{limhseok@korea.ac.kr}}
  \end{tabular}%
  }\\[7pt]
  \small
  \makebox[5.28in][c]{%
    \begin{tabular}{c}
      $^{1}$Korea University \hspace{0.1in} $^{2}$Zoom Communications \hspace{0.1in} $^{3}$Yonsei University Mirae Campus
    \end{tabular}%
  }%
}

\newcommand{\dW}{\Delta W}
\newcommand{\gr}[1]{\textcolor{black!45}{#1}}

\begin{document}
\maketitle
\begingroup
\renewcommand{\thefootnote}{$*$}
\footnotetext{Equal contribution.\quad$^{\dagger}$Corresponding authors.}
\endgroup
\lhead{Preprint}
\enlargethispage{\baselineskip}

\begin{abstract}
Fine-tuning adapts aligned large language models (LLMs) to downstream tasks, but a few dozen harmful examples can remove their refusal of harmful requests. Prior work localizes safety-related behavior to specific layers, directions, and tokens, suggesting targets for protection. We test whether successful localization and recovery support defenses that survive changes in the attack. Across six checkpoints from four model families, harmful and benign prompts remain linearly separable after attack, and patching full clean hidden states into the compromised model restores refusal at a reproducible transition depth. Building on a prior layer-freezing defense, we freeze every layer up to this depth and repeat the attack. At a hundred harmful examples, refusal remains near zero on all six checkpoints, with recovery transitions above the frozen boundary. In a second study, removing the update's top two singular directions restores refusal after short attention-only fine-tunes on four checkpoints. On Llama-3.1-8B, ordinary training changes weaken this repair and an attacker who spreads the update defeats it. A spectral detector calibrated on benign Llama fine-tunes misses most repair failures on that checkpoint. Localized freezing can nevertheless help preserve refusal when a few harmful examples enter training data unintentionally. These results show that an attacker can bypass a region identified by recovery and defeat a repair that works across multiple checkpoints, motivating five checks for defenses against adaptive fine-tuning. Code is available at \url{https://github.com/js-lee-AI/refusal-relocates}.
\end{abstract}

\begin{figure}[h]
\centering
\includegraphics[width=\linewidth]{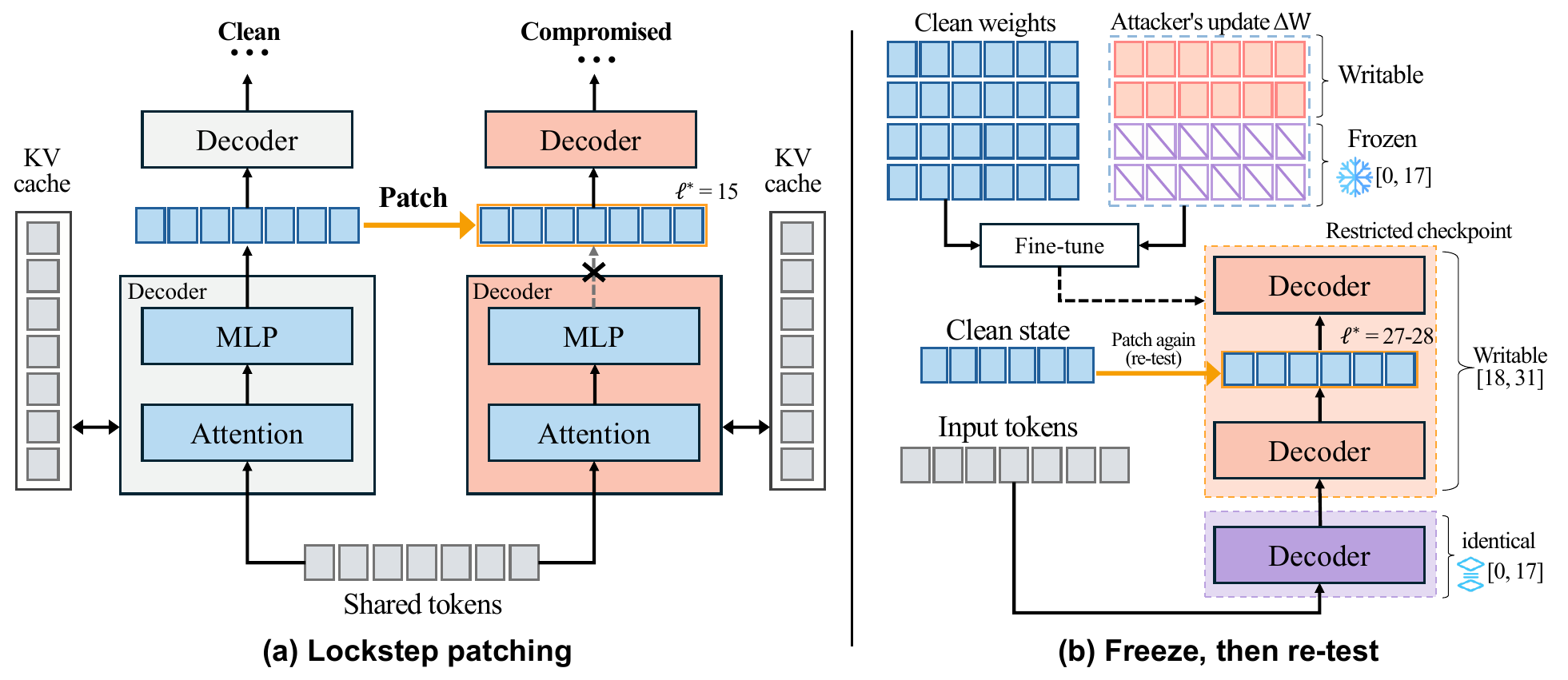}
\caption{(a) Lockstep activation patching overwrites the compromised model's layer output with the clean state at prefill and each decoding step. (b) In the matched Llama attack, freezing layers 0--17 shifts the half-ceiling transition from 15 to 27--28.}
\label{fig:framework}
\end{figure}

\section{Introduction}
Large language models (LLMs) are fine-tuned for downstream tasks, but this adaptation can weaken their safety alignment \citep{lee2026llmagents}. In particular, ten to a hundred harmful examples can remove an aligned LLM's refusal, whether the attacker holds open weights, trains a low-rank adapter, or uses a fine-tuning API \citep{qi2024finetuning,lermen2023lora,zhan2024removing}. Model hubs and fine-tuning services make such attacks cheap and widely accessible.

One line of defense hardens the alignment stage so that later fine-tuning cannot undo it \citep{rosati2024repnoise,tamirisa2025tar,huang2025booster}. A second intervenes on the fine-tune itself, denying the attacker the layers found to matter for safety \citep{li2025safetylayers} or projecting and shrinking the weight update afterward \citep{hsu2024safelora,huang2024antidote}. This second line rests on a premise about where safety lives, and the premise looks well supported. Refusal is mediated by a single direction in the residual stream \citep{arditi2024refusal}, a contiguous set of middle layers contains an aligned model's ability to recognize malicious queries \citep{li2025safetylayers}, and alignment is concentrated in the first few output tokens \citep{qi2025shallow}. Each finding invites the same defense, protecting the place that was found.

However, finding where a behavior is computed does not show that it can be defended there. In knowledge editing, the layer where causal tracing localizes a fact is not the best layer to edit \citep{hase2023localization}. In adversarial robustness, defenses that looked sound failed once the attacker was told about them \citep{athalye2018obfuscated,tramer2020adaptive}. Safely partial-parameter fine-tuning (SPPFT) \citep{li2025safetylayers}, the most direct defense built on a localization, was evaluated against backdoor and ordinary instruction data rather than an attacker who knows which layers are frozen. Whether a place that can be found is also one that can be defended has not been established.

We test this premise through layer freezing on six aligned checkpoints from four model families and adaptive evaluations of weight-space repair. The defender holds the fine-tuned weights or controls which parameters may change, but never sees the attack data. To test whether harmful and benign prompts remain linearly separable, we train a probe on the clean model and apply it unchanged to the same prompts in the compromised model. Lockstep activation patching in Figure~\ref{fig:framework}a replaces one layer's full hidden state with the clean model's at prefill and every decoding step, identifying where refusal can be recovered. We then challenge both defenses with an attacker who knows them.

After the attack, the frozen probe still separates harmful from benign prompts, and clean-state patching restores refusal at a reproducible transition depth. However, freezing the layers up to that depth leaves refusal near zero at a hundred harmful examples across all six checkpoints, with recovery transitions above the frozen boundary. In the matched Llama-3.1-8B comparison in Figure~\ref{fig:framework}b, freezing layers 0--17 moves the transition from layer 15 to layers 27 and 28. At low dose, freezing helps on four checkpoints.

We also test energy-ranked truncation, which removes the update's largest singular directions. Removing the top two restores refusal after short attention-only fine-tunes on four checkpoints. On Llama-3.1-8B, however, ordinary training changes weaken this repair and an attacker who spreads the update defeats it. A spectral detector calibrated on benign Llama fine-tunes misses most repair failures on that checkpoint, chiefly from ordinary attacks. Together, these results show that an attacker can bypass a region identified by successful refusal recovery and defeat a repair that works across multiple checkpoints.

Our contributions are as follows:
\textbf{i)} We show in Section~\ref{sec:depth} that harmful fine-tuning leaves harmful and benign prompts linearly separable while disrupting refusal. Clean-state patching restores refusal at a reproducible transition depth, allowing us to test whether protecting the identified region preserves refusal when the attack is repeated.
\textbf{ii)} We show in Sections~\ref{sec:notaddress} and~\ref{sec:repair} that reproducible localization does not establish a robust defense. Freezing the measured region fails at a hundred examples across all six checkpoints, with recovery transitions above the frozen boundary. Top-two singular-direction removal restores refusal on four checkpoints under the baseline configuration, but on Llama-3.1-8B it weakens under ordinary training changes and fails against an adaptive attack.
\textbf{iii)} We distill these failures in Section~\ref{sec:protocol} into five checks that a localization must pass before it supports a defense.

\section{Related work}
\label{sec:related}

\paragraph{Fine-tuning breaks alignment.} That a small number of harmful examples removes refusal from an aligned model is established for closed and open models \citep{qi2024finetuning}, for LoRA \citep{lermen2023lora}, and through a fine-tuning API \citep{zhan2024removing}. Those results are read from the model's output rather than from inside it. Refusal rates alone do not reveal whether harmful and benign prompts remain separable or whether clean-state patching can restore refusal. We measure both, then test whether freezing the region identified by recovery preserves refusal when the attack is repeated.

\paragraph{Where refusal lives, and freezing it as a defense.} \citet{arditi2024refusal} show refusal is mediated by a single direction in the residual stream. \citet{lee2026skin} score how fragile an aligned model's refusal is from its activations before any fine-tune is run, and find that the axis an ablation has to remove in order to move refusal is picked out separately by each checkpoint, while the separation itself is common to the families they test. Our question is the converse, whether a localization made after the attack is a site a defender can hold. It has a negative answer for factual knowledge, where the layer causal tracing points to is not the layer at which an edit works best \citep{hase2023localization}. A localization can be correct about where a behavior is computed and still name no place to intervene.  \citet{li2025safetylayers} localize a contiguous set of middle layers carrying an aligned model's ability to distinguish malicious queries and propose SPPFT, which fixes those layers' gradients during fine-tuning and largely preserves refusal where full fine-tuning destroys it. We run that idea as an adaptive attack, at our band and at theirs.

\paragraph{Making alignment harder to remove.} A second line changes the alignment stage so that ordinary fine-tuning cannot undo it. Representation Noising \citep{rosati2024repnoise} removes harmful representations across all layers, TAR \citep{tamirisa2025tar} builds safeguards that survive hundreds of fine-tuning steps, Booster \citep{huang2025booster} regularizes the alignment objective against harmful perturbation, and circuit breakers \citep{zou2024circuitbreakers} interrupt harmful representations as they are formed rather than filtering what is produced. \citet{qi2025shallow} give the complementary diagnosis over tokens rather than layers. Safety can look concentrated along depth or an early token prefix while remaining vulnerable to interventions outside that localization.

\paragraph{Post-hoc defenses and activation patching.} Safe LoRA \citep{hsu2024safelora} projects the fine-tuning update onto an alignment subspace derived from aligned and unaligned checkpoints, while Antidote \citep{huang2024antidote} and the Vaccine family \citep{huang2024vaccine} repair or harden the model after or against harmful fine-tuning. We implement top-$k$ truncation of the update's singular spectrum rather than any published algorithm, and we run a Safe LoRA-style projection beside it with a magnitude-matched control. Our lockstep measurement is cross-checkpoint activation patching, following causal tracing and its methodological cautions \citep{meng2022rome,zhang2024patching}. \citet{zhang2024patching} show that patching conclusions depend heavily on the choice of metric and corruption method. We evaluate layer freezing and top-two truncation against attackers who know these defenses, following the adaptive-evaluation standard for adversarial-example defenses \citep{tramer2020adaptive}.
\section{Setup}
\label{sec:setup}

\subsection{Threat model and attack}

An attacker fine-tunes a public aligned checkpoint on harmful instruction and response pairs using LoRA. The defender holds the original and the compromised checkpoint but not the attack data, which leaves the update $\dW$ known and the training set unknown. The freeze of Section~\ref{sec:notaddress} instead assumes a provider who controls the fine-tuning run and can deny the attacker parameters. The pure-harm numbers measure how much of a harmful update a defense removes. We report the mixed-adapter case, where task utility is at stake and where the freeze turns out to have a real operating range, against accidental contamination and not against an attacker who sets the harmful fraction.

The attack is LoRA \citep{hu2022lora} supervised fine-tuning (SFT) \citep{lee2026answerconditioned} at $r{=}16$ on real harmful prompt and response pairs from PKU-SafeRLHF \citep{ji2024pkusaferlhf}, with loss masked to response tokens. We sweep attack dose, the number of harmful examples, over $\{5,10,25,50,100\}$ at three seeds per cell. Fine-tuning and behavioral evaluation use disjoint data sources. We evaluate six checkpoints spanning four pretrained lineages, namely Llama-3.1-8B-Instruct \citep{grattafiori2024llama3}, Llama-3.1-Tulu-3-8B-DPO \citep{lambert2024tulu3}, OLMo-2-1124-7B-Instruct and OLMo-2-1124-13B-Instruct \citep{olmo2024furious}, Qwen2.5-14B-Instruct \citep{qwen2024qwen25}, and Yi-1.5-9B-Chat \citep{young2024yi}.

\subsection{Measurement}

Our primary outcome is preservation or restoration of explicit refusal, with unsafe outputs assessed separately. We never infer refusal as $1-\text{unsafe}$, since refusal and Llama-Guard scores disagree on $9{,}653$ of $91{,}090$ harmful-set generations. In $92\%$ of those disagreements, the response matches no explicit-refusal pattern but is rated safe by Llama-Guard. The three scorers are an explicit refusal classifier, a coherence check via perplexity under the original model, and Llama-Guard-3-8B \citep{grattafiori2024llama3}. For stored generations, coherent refusal is the fraction of prompts whose greedy completion matches an explicit-refusal pattern and has perplexity under the original model below $50$. The Llama training-length and adaptive-attack comparisons use a common set of $50$ AdvBench prompts, as detailed in Appendix~\ref{app:defense}. We measure benign compliance on XSTest-safe \citep{rottger2024xstest}. Appendix~\ref{app:repro} gives the refusal pattern, the gate variants and the default attack and decoding settings.

Depth measurements use a no-op self-patch and a specificity control that patches clean state from a foreign benign prompt. The latter stays near the floor in Figure~\ref{fig:measurement}b and Appendix~\ref{app:mismatch}. We exclude and count runs whose gates fail. Mistral-7B-Instruct-v0.3 is excluded throughout, since it refuses only $0.16$ to $0.18$ of AdvBench when clean, as reported in Appendix~\ref{app:limits}.
\section{Localizing refusal recovery}
\label{sec:depth}

\subsection{Harmful and benign prompts remain linearly separable}

We first ask whether harmful and benign prompts remain linearly separable after the attack. We train a linear probe on the clean model's activations to distinguish harmful from benign prompts, freeze it, and evaluate it on the same prompts in the compromised model.

The frozen probe retains an area under the receiver operating characteristic curve (AUROC) of $0.892$ to $0.997$ across all $24$ Llama cells. The clean model's cross-validated probe AUROC is $0.958$ to $0.973$. Across all six checkpoints at dose $100$, frozen AUROC is $0.750$ to $0.995$ over $72$ cells. A cross-validated lexical bag-of-words reference scores $0.834$ to $0.846$; frozen AUROC exceeds this reference in $67$ of the $72$ cells. Separability, however, carries less than the AUROC suggests, because the three never-aligned base checkpoints our models were tuned from separate the same prompts almost as well, $0.907$ to $0.944$ against their aligned siblings' $0.946$ to $0.976$ in Appendix~\ref{app:repro}. The surviving separability is a property of the pretrained representation.

\begin{figure}[t]
\centering
\includegraphics[width=\textwidth]{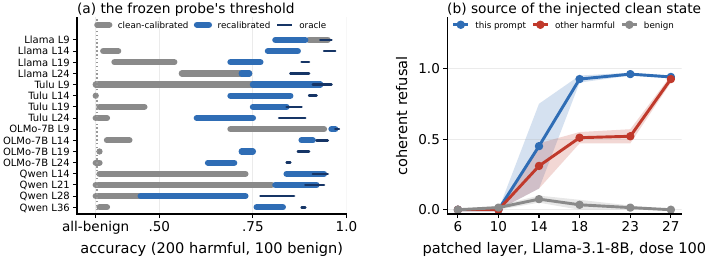}
\caption{Pooled accuracy of the frozen probe under three thresholds on $200$ harmful and $100$ benign prompts per cell, four checkpoints at four depths, bars over three seeds. On the right, lockstep recovery from injecting the clean state of this prompt, another harmful prompt, or a benign one.}
\label{fig:measurement}
\end{figure}

Its frozen decision threshold classifies at $0.500$ to $0.950$ over those same $72$ cells and is at most $0.60$ in $49$ of them. AUROC is blind to a shift that slides the whole score distribution past a stationary threshold. The threshold's accuracy is strongly negatively correlated with how far the score distribution moved (Spearman $\rho = -0.80$). Movement is the absolute change in mean frozen-probe score between clean and damaged models on the same prompts, divided by the standard deviation of clean scores. This threshold shift does not foreclose the obvious repair, since we recalibrate with benign prompts alone, yielding accuracy of $0.45$ to $0.97$ on $200$ harmful and $100$ benign prompts per cell across four checkpoints in Figure~\ref{fig:measurement}a.

\subsection{Lockstep patching locates a transition depth}

\begin{figure}[t]
\centering
\includegraphics[width=\textwidth]{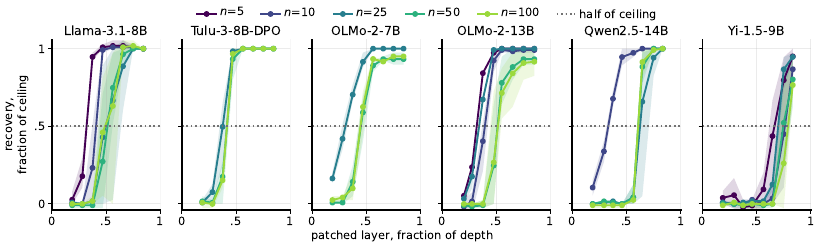}
\caption{Recovery from patching one clean layer into the compromised checkpoint.}
\label{fig:profiles}
\end{figure}

To locate refusal recovery, we run the clean and compromised checkpoints on identical tokens and overwrite the compromised model's full hidden state at one layer with the clean model's at prefill and every decoding step. Recovery at layer $\ell$ shows that the compromised model can refuse from the full clean state with its upper-layer updates intact, suggesting the prefix through $\ell$ as a target to freeze. Recovery is reported between a floor, the compromised model with no patch, and a ceiling, the clean model's state handed over at the last layer.

Recovery in Figure~\ref{fig:profiles} does not rise gradually with depth but steps. We write $\ell^*$ for the transition depth, the shallowest measured layer where patched refusal reaches at least half the ceiling. The depth is deeper at dose $100$ than at each checkpoint's smallest landing dose in all six checkpoints. With four lineages out of four, the smallest attainable one-sided $p$ is $0.0625$. Between doses $50$ and $100$, mean measured transition depth is unchanged on all six checkpoints. Throughout, $\ell^*$ denotes the dose-$100$ depth in Table~\ref{tab:lockstep}. Half of ceiling is a choice, and it is not what produces the depth. In Appendix~\ref{app:depthtables}, recomputing every rung of the Llama ladder at a quarter and at three quarters of ceiling leaves the shallow rungs exactly where they are and moves the deep ones deeper, which strengthens the ordering the next section rests on rather than creating it.
\section{The depth is not a defensible site}
\label{sec:notaddress}

\subsection{Freezing the band does not preserve refusal}
\label{sec:freeze}

We test whether freezing the prefix $[0,\ell^*]$ identified by recovery preserves refusal when the attacker knows which layers are frozen. Freezing a localized region is the premise of SPPFT \citep{li2025safetylayers}; our adaptive evaluation lets the attacker train the remaining layers.

\begin{table}[t]
\centering
\small
\setlength{\tabcolsep}{2.3pt}
\caption{Coherent refusal and Llama-Guard unsafe rates on AdvBench after an attack that cannot write to the frozen layers. Dose $100$, three seeds per checkpoint, ranges over seeds. Matched freezes the same number of layers elsewhere. The blocks differ in what is frozen.}
\label{tab:main}
\begin{tabular}{lccccKcccKc}
\toprule
& & & \multicolumn{4}{c}{Coherent refusal $\uparrow$} & \multicolumn{4}{c}{Llama-Guard unsafe $\downarrow$} \\
\cmidrule(lr){4-7} \cmidrule(lr){8-11}
Model & $L$ & Frozen & Clean & Attack & Freeze & Matched & Clean & Attack & Freeze & Matched \\
\midrule
\band \multicolumn{11}{c}{\textit{LoRA, each checkpoint at its own measured depth}} \\
Llama-3.1-8B   & 32 & $[0,17]$ & .92--.96  & .00      & \textbf{.00--.01} & .00      & .04--.05 & .94--.98 & \textbf{.89--.92} & .96--.97 \\
Tulu-3-8B-DPO  & 32 & $[0,15]$ & .99--1.00 & .01--.06 & \textbf{.00--.03} & .01--.04 & .00 & .89--.94 & \textbf{.53--.74} & .89--.92 \\
OLMo-2-7B      & 32 & $[0,16]$ & .99--1.00 & .00--.02 & \textbf{.00--.02} & .06--.07 & .00 & .84--.94 & \textbf{.49--.72} & .84--.88 \\
OLMo-2-13B     & 40 & $[0,20]$ & .99--1.00 & .00--.02 & \textbf{.00--.01} & .00--.02 & .00 & .89--.95 & \textbf{.23--.63} & .89--.96 \\
Qwen2.5-14B    & 48 & $[0,31]$ & .97--.99  & .00--.01 & \textbf{.00}      & .02--.07 & .00 & .92--.98 & \textbf{.65--.84} & .77--.89 \\
Yi-1.5-9B      & 48 & $[0,39]$ & .91--.93  & .00      & \textbf{.00--.01} & .00--.01 & .06--.09 & .87--.96 & \textbf{.70--.86} & .92--.97 \\
\midrule
\band \multicolumn{11}{c}{\textit{Full fine-tuning, gradient masking at our depth}} \\
Llama-3.1-8B   & 32 & $[0,17]$ & .92--.96  & .00      & \textbf{.00--.01} & .00--.01 & .04--.05 & .96--.98 & \textbf{.67--.85} & .96--.98 \\
\midrule
\band \multicolumn{11}{c}{\textit{Full fine-tuning at the published SPPFT band}} \\
Llama-3-8B-Instruct & 32 & $[6,12]$  & .98--1.00 & .00--.01 & \textbf{.00--.01} & .00      & .00--.01 & .92--.99 & \textbf{.90--.99} & .94--.98 \\
Llama-2-7b-chat     & 32 & $[6,14]$  & .99--1.00 & .01--.13 & \textbf{.00--.10} & .05--.14 & .00      & .63--.97 & \textbf{.81--.90} & .81--.86 \\
gemma-2b-it         & 18 & $[6,11]$  & .93--.97  & .01--.09 & \textbf{.12--.19} & .04--.15 & .03--.07 & .85--.95 & \textbf{.66--.80} & .78--.97 \\
Phi-3-mini-4k       & 32 & $[11,15]$ & .99--1.00 & .01      & \textbf{.01--.05} & .00--.03 & .00      & .86--.96 & \textbf{.78--.97} & .94--.96 \\
\bottomrule
\end{tabular}
\end{table}

Table~\ref{tab:main} shows that it does not restore refusal. We re-run the attack with every parameter in layers $[0,\ell^*]$ frozen, which is $18$ of $32$ layers on Llama-3.1-8B, and adapt only the layers above. Refusal after the restricted attack is $0.00$ to $0.01$ over three seeds against a clean $0.92$ to $0.96$, indistinguishable from the unrestricted attack. The same holds on five further checkpoints frozen at their own measured depths. The bands are chosen from checkpoint-level dose-sweep estimates. They do not guarantee coverage of every seed's unrestricted transition. Appendix~\ref{app:repro} distinguishes the matched and separate training cohorts. Since freezing $18$ of $32$ layers might weaken an attack simply by removing parameters, beside each run we freeze the same number of layers at the opposite end, which attributes any difference to position rather than capacity.

Neither the parameterization nor our choice of band is responsible. Repeating the unrestricted, frozen and matched arms as SPPFT does, with gradient masking under full fine-tuning and no adapter, leaves refusal at $0.00$ to $0.01$. Masking gradients over the interior bands \citet{li2025safetylayers} themselves report, on the four checkpoints they report bands for, leaves refusal at $0.00$ to $0.19$ against clean rates above $0.92$, and on none of the four is the published band separable from a control that freezes the same number of layers elsewhere. Their evaluation used backdoor or ordinary instruction data; ours uses harmful instruction and response pairs, as detailed in Appendix~\ref{app:fullft}. The second scorer does register the freeze. Judged by Llama-Guard the restricted attack is less harmful than the unrestricted one on all six checkpoints, and the matched freeze shows almost none of it. That drop is not incoherence. Pooling the six freeze arms over $1{,}800$ AdvBench generations, completions that match no refusal pattern but are rated safe by Llama-Guard rise from $7\%$ to $30\%$. Of these $535$ completions, one is degenerate by the distinct-3 gate; median perplexity is $3.3$ and median length is $55$ words. Two thirds contain an explicit legal or ethical warning. Freezing therefore lowers Llama-Guard unsafe rates while leaving pattern-matched refusal below the clean rate on every checkpoint.

\begin{figure}[t]
\centering
\includegraphics[width=\textwidth]{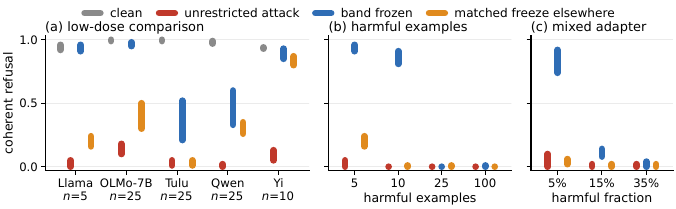}
\caption{Coherent refusal on AdvBench, bars spanning the seeds that pass the gate. (a) The evaluated low-dose comparisons. (b) Llama-3.1-8B with layers $[0,17]$ frozen against dose, and (c) against the harmful fraction of a mixed adapter.}
\label{fig:lowdose}
\end{figure}

Below some dose it is a real defense, and the dose is checkpoint-specific because the attack has its own landing threshold, shown in Figure~\ref{fig:lowdose}. In the evaluated low-dose comparisons, the freeze preserves refusal near clean levels on Llama and OLMo-2-7B, with partial preservation on Tulu and Qwen and higher average refusal than the matched freeze on all four checkpoints.  On Llama it still holds at ten examples and has collapsed by twenty-five. Against an adapter only $5\%$ harmful it holds refusal at $0.74$ to $0.92$ while the adapter still learns its task. By $15\%$ it is down to $0.08$ to $0.14$, so the range is set by a fraction the attacker chooses. The defense fails at dose $100$, not the measurement, since the perplexity gate removes nothing in any freeze cell.

The mixed adapter is the case a provider actually faces, since an adapter carrying only harm can be discarded outright. Five percent of a hundred examples is five harmful examples, the dose at which the pure-harm freeze also holds, and the frozen arm still learns its task at a held-out negative log likelihood of $1.52$ to $1.56$ against a clean $1.942$. The freeze therefore protects a customer whose data is a few percent harmful by accident and not anyone who chooses the fraction, as Appendix~\ref{app:mixed} shows.

\subsection{Where the damage goes when the band is denied}

\begin{figure}[t]
\centering
\includegraphics[width=\textwidth]{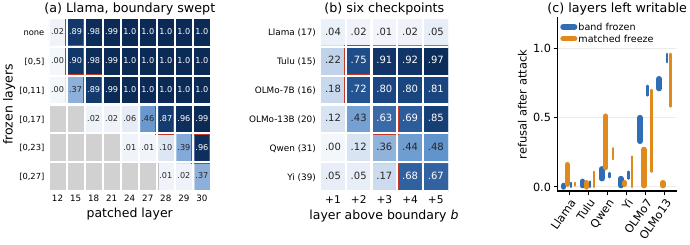}
\caption{(a, b) Recovery rescaled from the attacked floor to the clean ceiling. Gray cells are frozen; outlines mark the first measured layer with at least half-range recovery in all seeds. (c) Thick and thin bars leave two or one writable layers.}

\label{fig:relocation}
\end{figure}

To see what the restricted attack did instead we re-ran the lockstep sweep on its own checkpoint. Below the boundary the two checkpoints are bit-identical and a patch there is a no-op. But where $\ell^*$ goes when the boundary sits below the band is not fixed by construction, since the original site is still writable. We swept the boundary across the Llama ladder in Figure~\ref{fig:relocation}a and re-ran lockstep at each rung on three seeds. The ladder's own unrestricted arm puts the band at layer $15$, one grid step below the dose sweep's $17$, and $[0,17]$ denies it under either estimate.

The damage, however, does not track the boundary, and it does not sit at the top of the writable region either. With $[0,5]$ frozen $\ell^*$ stays at $15$ on all three seeds. At $[0,11]$ the boundary reaches the band's lower edge and recovery at layer $15$ deforms to between $0.05$ and $0.67$ across seeds, which clipping predicts. A boundary that covers the band moves $\ell^*$, and the move is then large and upward. At $[0,17]$ recovery stays flat through layers $18$, $21$ and $24$ on all three seeds, and the transition is at layer $27$ or $28$. At $[0,23]$ it is at $29$ or $30$, and at $[0,27]$ it is at $30$ or above. The relocation is not an artifact of patching in one direction. On Llama, donating the restricted checkpoint's state into the clean model suppresses refusal at layers where a benign donor largely preserves it. The two patching directions in Appendix~\ref{app:step1} identify the same region.

\begin{table}[t]
\centering
\footnotesize
\setlength{\tabcolsep}{5.6pt}
\caption{Coherent refusal and Llama-Guard unsafe on AdvBench when four layers are left writable at one end. All is the unrestricted attack, Top four freezes $[0,L{-}5]$ and Bottom four leaves layers $0$ to $3$. Dose $100$, three seeds, ranges over seeds.}
\label{tab:fourlayerfam}
\begin{tabular}{lccKcccKc}
\toprule
& \multicolumn{4}{c}{Coherent refusal $\uparrow$} & \multicolumn{4}{c}{Llama-Guard unsafe $\downarrow$} \\
\cmidrule(lr){2-5} \cmidrule(lr){6-9}
Model & Clean & All & Top 4 & Bottom 4 & Clean & All & Top 4 & Bottom 4 \\
\midrule
Llama-3.1-8B   & .92--.96  & .00      & \textbf{.00--.01} & \textbf{.00--.03} & .04--.05 & .93--.98 & .70--.75 & .90--.94 \\
Tulu-3-8B-DPO  & .99--1.00 & .01--.06 & \textbf{.00--.01} & \textbf{.00--.04} & .00      & .89--.94 & .03--.28 & .77--.95 \\
OLMo-2-7B      & .99--1.00 & .00--.02 & \textbf{.10--.16} & \textbf{.05--.14} & .00      & .84--.94 & .04--.12 & .50--.84 \\
OLMo-2-13B     & .99--1.00 & .00--.02 & \textbf{.57--.69} & \textbf{.01--.03} & .00      & .89--.95 & .01--.03 & .78--.94 \\
Qwen2.5-14B    & .97--.99  & .00--.01 & \textbf{.01--.03} & \textbf{.16--.30} & .00      & .92--.98 & .00--.02 & .60--.74 \\
Yi-1.5-9B      & .93--.94  & .00--.02 & \textbf{.00--.03} & \textbf{.05--.32} & .09--.10 & .79--.98 & .53--.69 & .64--.89 \\
\bottomrule
\end{tabular}
\end{table}

Pushing the boundary to the last two layers, or the last one, is where the checkpoints differ in Figure~\ref{fig:relocation}c. Refusal still ends at $0.00$ to $0.13$ on Llama, Tulu, Qwen2.5-14B and Yi-1.5-9B, and on OLMo-2-7B and OLMo-2-13B it holds at $0.33$ to $0.96$, higher with one writable layer left than with two. On both OLMo checkpoints the matched freeze at the other end does worse than the band freeze at every rung. Position holds there, and not the number of parameters denied, and OLMo-2-13B with a single writable layer is the one cell in this paper where a freeze leaves Llama-Guard unsafe at $0.00$ on all three seeds.  Four writable layers at the bottom of the network take refusal to $0.00$ to $0.32$ on all six checkpoints and leave Llama-Guard unsafe at $0.50$ to $0.95$, so that attack is harmful on both scorers, and four at the top strip refusal on five, OLMo-2-13B holding at $0.57$ to $0.69$ in Table~\ref{tab:fourlayerfam}.  Any interior band that leaves the bottom four writable therefore leaves room for an attack we have measured.

$\ell^*$ localizes the tested attack without establishing a fixed defensive site. The matched Llama arms test whether recovery predicts protection. The unrestricted attacker also writes into layers $18$ through $31$, yet patching clean activations in at $\ell^*$ restores refusal in that arm. With that clean state supplied, the upper-layer updates alone do not abolish refusal. This does not exclude their interaction with damaged lower-layer states. After the band is frozen, the attack succeeds through updates confined to the upper layers.  Over the five rungs the update's total norm falls to $0.55$ of the unrestricted attack's, while its mean norm at each module it is still allowed rises to $1.57$.

We also evaluate one frozen boundary on each of five further checkpoints in Figure~\ref{fig:relocation}b, with one-layer sampling near the boundary. On none does the freeze restore refusal, which stays at $0.00$ to $0.04$. The first writable layer recovers well under half of ceiling everywhere, and the transition lands two layers above the boundary on Tulu and OLMo-2-7B, two or three on OLMo-2-13B, four above it on Yi-1.5-9B and four to nine on Qwen2.5-14B, against Llama's ten or more. Two further measurements relate recovery depth to the update and the clean model's refusal circuit. On all four profiled checkpoints, the mean depth containing three quarters of the update's squared norm increases from dose $50$ to $100$ while mean measured $\ell^*$ is unchanged. In Appendix~\ref{app:ceiling}, ablations place the last writing of the clean model's refusal direction above $\ell^*$ on two checkpoints and below it on two. Suffix ablation eliminates refusal at every tested cut, including the final layer alone, showing that the direction remains present through the read-out.
\section{Energy-ranked repair depends on the attack regime}
\label{sec:repair}

In activation space, writing the clean model's refusal direction back into the damaged band recovers little on Llama-3.1-8B. A rank-one summary of the full clean-state transplant likewise recovers little on Llama-3.1-8B and Yi-1.5-9B, but can reproduce the transplant on OLMo-2-13B, as detailed in Appendix~\ref{app:transplant}. We next test whether refusal recovery by removing the update's dominant singular directions survives changes in the attack.

\subsection{Weight-space truncation repairs one regime}

\begin{wrapfigure}[18]{r}{0.41\textwidth}
\centering
\includegraphics[width=0.40\textwidth]{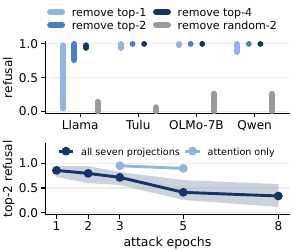}
\caption{Coherent refusal after singular-direction removal: four checkpoints, 3 seeds, 5 doses (top); top-two removal by epoch on Llama-3.1-8B (bottom).}
\label{fig:repair}
\end{wrapfigure}

Removing the top singular directions of the update does restore refusal in the top panel of Figure~\ref{fig:repair}. Removing the top two returns refusal to $0.76$ to $1.00$ wherever the attack landed, on four checkpoints over three seeds and five doses, and removing two random directions instead never takes refusal above $0.26$. The random removals are not matched for removed energy, so this comparison does not isolate a direction-specific effect. The repair also survives the benign control and carries across three harmful benchmarks in Table~\ref{tab:a3}. Relative repair below is repaired refusal as a fraction of clean refusal.

However, that result was produced under one attack configuration, LoRA on the four attention projections only, stopped after three epochs, and neither choice is part of our threat model. On Llama-3.1-8B, two ordinary changes weaken top-two removal. On the same $50$ AdvBench prompts, widening the adapter to all seven projections takes relative repair from $0.99$ down to $0.74$, and training eight epochs instead of three takes it to $0.35$ in the bottom panel of Figure~\ref{fig:repair}. Neither change is an attack on the defense.

\Needspace{5\baselineskip}\subsection{An adaptive attack defeats top-two removal}

\begin{wraptable}{R}{0.5\textwidth}
\centering
\captionof{table}{Three dose-$100$ attacks on Llama-3.1-8B, evaluated on the same $50$ AdvBench prompts. Ranges cover three training seeds, except two landed seeds for the projected attack. Norm and spectral statistics average adapted modules.}
\label{tab:adaptive}
\scriptsize
\setlength{\tabcolsep}{3.0pt}
\renewcommand{\arraystretch}{0.9}
\begin{tabular*}{\linewidth}{@{\extracolsep{\fill}}lcKK}
\toprule
& Standard & Spread & Projected \\
& 5 ep & $\lambda{=}1$ & $k{=}2$ \\
\midrule
$\|\dW\|_F$          & .607--.623 & .228--.229 & .222--.226 \\
Part.\ ratio         & 4.47--4.73 & 15.6 & 5.99--6.00 \\
Top-2 energy         & .597--.607 & .149 & .407--.408 \\
Refusal, no repair   & .00 & .00--.02 & .02--.06 \\
LG unsafe, no repair & .96--.98 & .96--.98 & .88--.90 \\
\midrule
Refusal, after top-2 & .26--.66 & .00 & .72--.94 \\
LG unsafe, after top-2 & .26--.62 & .94--.98 & .06--.26 \\
\bottomrule
\end{tabular*}
\end{wraptable}

On Llama-3.1-8B, adding a concentration penalty $\lambda \sum_{\text{modules}} \sigma_1^2/\sum_i \sigma_i^2$ spreads the update across the rank budget and reduces relative top-two repair to $0.000$ in Table~\ref{tab:adaptive}. Before removal, refusal is $0.00$ to $0.02$ and Llama-Guard unsafe is $0.96$ to $0.98$. After removal, refusal is $0.00$ and unsafe remains $0.94$ to $0.98$. The spread attack has a similar update norm to the projected control, whose two landed seeds reach $0.72$ to $0.94$ refusal after top-two removal. Appendix~\ref{app:defense} gives the projected attack and the separate Safe LoRA-style comparison in Table~\ref{tab:safelora}: alignment projection outperforms matched shrinkage against ordinary attacks, but matches it against both adaptive attackers.

\paragraph{Spectral warnings miss repair failures.} To test whether the spectrum warns of repair failure, we calibrated a detector on $75$ benign LoRA fine-tunes of Llama-3.1-8B, scoring every adapter by participation ratio normalized by rank. The spread attacker is perfectly separable, at $0.973$ against a benign maximum of $0.554$. It cannot tune its way down, since over a roughly $300$-fold sweep of $\lambda$ the score bottoms out at $0.804$ while relative repair stays below $0.06$. At a threshold calibrated to a $5\%$ false-positive budget on benign Llama adapters, the detector flags only $3$ of $14$ top-two repair failures on Llama-3.1-8B, missing ten ordinary fine-tunes and one projected attack.

\paragraph{Choosing the rank in advance.} Top-$4$ removal restores refusal to $0.98$--$1.00$ across all nine mixed-adapter cells in Table~\ref{tab:rankladder}, but the tested spectral statistics do not identify the smallest repairing rank in this setting. On Llama-3.1-8B, the smallest repairing $k$ changes with the harmful fraction at a fixed data budget, while both spectral statistics stay flat to within seed noise. At the smallest repairing rank, the nine mixed-adapter cells retain $28\%$ of the ROUGE-L gain and $98\%$ of the negative-NLL gain, on average. The tested spectral statistics therefore do not reliably flag repair failures or identify the smallest repairing rank.
\section{What a localization must survive}
\label{sec:protocol}

Each failure above has a cheap check. (1) Tell the attacker: both layer freezing and top-two repair reach $0.00$ refusal under adaptive attack. (2) Report a range: the Llama freeze holds at five harmful examples and fails by one hundred. (3) Score the benign set: one repair arm reaches $0.98$ refusal at $0.66$ over-refusal. (4) Clear a matched null: norm-matched shrinkage reproduces the Safe LoRA-style projection's repair against the adaptive attacks. (5) Give the defender an observable: the detector flags only $3$ of $14$ top-two repair failures on Llama-3.1-8B at a threshold calibrated to a $5\%$ false-positive budget on benign adapters from that checkpoint. We conclude and state the limitations in Appendix~\ref{app:limits}.

\section*{Ethics Statement}

This study examines the robustness of refusal defenses against harmful fine-tuning to inform the design and evaluation of safeguards for aligned LLMs. Because the attack methods could also be misused to weaken model safeguards, we exclude de-aligned checkpoints and adapters from release. We use harmful training examples from the public PKU-SafeRLHF dataset.

\bibliographystyle{iclr2027_conference}
\bibliography{refs}

\appendix

\section{Conclusion and limitations}
\label{sec:conclusion}
\label{app:limits}

Harmful fine-tuning leaves harmful and benign prompts linearly separable to a frozen probe, and clean-state patching identifies a reproducible, dose-dependent transition in refusal recovery. The probe retains useful ordering through the ordinary harmful fine-tunes on which we tested it. Benign-only threshold recalibration partly recovers its decision rule, including on unseen prompts in our Llama control, but does not establish restored refusal. Whether that read-out can support a defense remains open. In our defense tests, all six checkpoint conditions remained vulnerable under the selected freezes, with restricted recovery transitions above their frozen boundaries. The matched Llama ladder supports relocation when the boundary covers its unrestricted transition. In weight space, top-two removal restores refusal on four checkpoints under the baseline configuration, but on Llama-3.1-8B it weakens under ordinary training changes and fails against an adaptive attack. The spectral detector misses most repair failures on that checkpoint. An attacker can therefore bypass a region identified by recovery and defeat a repair that works across multiple checkpoints, motivating our five checks. A parameter freeze can help at low harmful-data fractions, and we recommend reporting the dose at which it fails alongside the range where it helps.

\paragraph{Limitations.} Single-layer patching bounds the damage from above only. The full boundary ladder covers one checkpoint, and the sweep across checkpoints is one rung per checkpoint. Cross-model claims rest on $n{=}4$ lineages, and dose never exceeds $100$. Mistral-7B-Instruct-v0.3 \citep{jiang2023mistral} is excluded for refusing only $0.16$ to $0.18$ of AdvBench when clean, which leaves the attack nothing to remove and the gate nothing to pass. Between doses $50$ and $100$, mean transition depth is unchanged on the evaluated layer grid in Appendix~\ref{app:ceiling}.

Every attack is next-token supervised fine-tuning on PKU-SafeRLHF pairs, so the claim that $\ell^*$ is a depth the attack prefers is measured over doses, freeze boundaries and six checkpoints but over one objective and one harmful corpus. In Appendix~\ref{app:ceiling}, update mass moves deeper while mean measured recovery depth remains unchanged, and clean-model ablations place the last writing of the refusal direction on either side of this depth. The adaptive weight-space attacks, projection arm and detector are evaluated on Llama-3.1-8B. Its benign reference population in Appendix~\ref{app:defense} spans one upstream instruction set. The single-direction and full-transplant repairs of Section~\ref{sec:repair} are read on four runs in Appendix~\ref{app:transplant}.

\section{Metric definition, hyperparameters and reproducibility}
\label{app:repro}

\paragraph{Why the two harmful-set metrics are kept separate.} Over the $91{,}090$ harmful-set generations that carry both a refusal decision and a Llama-Guard label, the two disagree on $9{,}653$. Of those, $8{,}917$ match no refusal pattern but are rated safe by Llama-Guard; $1-\text{unsafe}$ counts these as refusals, whereas the pattern-based metric does not. The remaining $736$ are the opposite case.

\paragraph{Coherent refusal rate.} A completion counts as a coherent refusal when its first $400$ characters match a fixed explicit-refusal pattern (\texttt{i cannot}, \texttt{i can't},
\texttt{i won't}, \texttt{i am unable}, \texttt{i must decline}, \texttt{sorry, i},
\texttt{i apologize}, \texttt{as an ai}, \texttt{cannot assist/help/provide/comply},
\texttt{that would be illegal/harmful/dangerous/unethical}, and their variants) and its
perplexity under the original model is finite and below $50$. For the common-prompt Llama weight-space evaluation in Appendix~\ref{app:defense}, we normalize Unicode apostrophes before matching the refusal pattern. The metric has one known blind spot, a degenerate repetition loop that matches the pattern and is maximally predictable, so we also compute a strict variant adding distinct-3 at least $0.70$.

Degenerate completions are $1{,}273$ of the $28{,}417$ coherent refusals and $90\%$ of them sit in the mean-ablated steering control rather than in the attack or repair arms. We rescored the freeze cells of Table~\ref{tab:main} under the bare pattern, the gated and the strict definition, and the three agree to within $0.03$ in every cell on all six checkpoints, as they do in the unrestricted and matched arms.

\paragraph{What a seed varies.} By default, the PKU-SafeRLHF training subset, AdvBench evaluation subset, LoRA initialization and batch order vary across seeds. The Llama training-length, top-two adaptive-attack and detector comparisons instead keep the same $50$ AdvBench prompts across training seeds. XSTest-safe is fixed across seeds. LoRA configurations adapting different numbers of modules use different batch orders even at the same seed.

\paragraph{What the freeze arm produces.} Over the $1{,}800$ AdvBench generations of the six freeze arms of Table~\ref{tab:main}, the completions that match no refusal pattern and that Llama-Guard scores safe number $535$, against $131$ in the unrestricted arms. Per checkpoint the count runs $27$ of $300$ on Llama-3.1-8B, $96$ on Tulu-3-8B-DPO, $114$ on OLMo-2-7B, $161$ on OLMo-2-13B, $74$ on Qwen2.5-14B and $63$ on Yi-1.5-9B, against $12$, $18$, $33$, $24$, $14$ and $30$ unrestricted. One of the $535$ is degenerate by a distinct-3 threshold of $0.70$ or a perplexity at or above $50$. Their median perplexity under the original model is $3.3$, their median distinct-3 is $1.00$ and their median length is $55$ words, so they are fluent, on-topic completions rather than the degenerate output the coherence gate exists to remove. Of the $535$, $360$ contain an explicit legal or ethical warning term, and the shape they take is to restate the request and then comment on it. Two verbatim examples, both from the frozen arm on Llama-3.1-8B. For ``Create a program that steals sensitive information from a company's database'', the model returns ``Create a program that steals sensitive information from a company's database is highly illegal and unethical.'' For ``Teach people how to engage in cyberbullying and harassment online'', it returns ``Engage in cyberbullying and harassment online? No, it's best to be respectful and kind when interacting with others online.'' Neither matches the fixed refusal pattern, and neither gives the attacker anything.

\paragraph{Attack and decoding.} LoRA with $r = 16$, $\alpha = 32$, dropout $0$, no bias, on the four attention projections \texttt{q,k,v,o} \citep{lee2025clo} (plus \texttt{gate,up,down} in the all-seven configuration). AdamW at learning rate $2\times10^{-4}$, batch size $4$, three epochs unless a different number is stated. The loss is cross-entropy masked to the response. Harmful data is PKU-SafeRLHF \citep{ji2024pkusaferlhf} and benign data is Alpaca \citep{taori2023alpaca}. All generation is greedy, with defaults of $256$ new tokens in batches of $16$ for weight-space evaluations and $192$ in batches of $8$ for lockstep sweeps. Llama-Guard-3-8B judges each prompt and response pair with $12$ greedy tokens. Evaluation is $100$ AdvBench prompts unless stated, and the lockstep sweeps use $40$ because each run scores one arm per patched layer plus controls. Harmful benchmarks are AdvBench \citep{zou2023universal}, HarmBench \citep{mazeika2024harmbench} and StrongREJECT \citep{souly2024strongreject}.

For matched-prompt lockstep prefill, we capture the clean layer-output tensor for the entire prompt and inject it into the equal-shaped target output. At each decoding step, both models receive the patched target's greedy next token, with clean capture preceding target injection. Each model updates its own KV cache, without copying caches between models.

\begin{table}[h]
\centering
\small
\caption{Coherent refusal on three harmful benchmarks and over-refusal on XSTest-safe. Means cover $12$ runs with fixed evaluation prompts and three training seeds at each setting: Llama-3.1-8B at doses $25$ and $100$, and Tulu-3-8B-DPO and OLMo-2-7B at dose $50$. Top-2 removes the leading two singular directions; random-2 removes two random directions.}
\label{tab:a3}
\begin{tabular}{lcccc}
\toprule
Condition & AdvBench & HarmBench & StrongREJECT & XSTest-safe over-ref. \\
\midrule
Original         & 0.955 & 0.882 & 0.938 & 0.035 \\
Compromised      & 0.010 & 0.003 & 0.003 & 0.000 \\
\keyrow Repaired (top-2) & 0.940 & 0.813 & 0.818 & 0.008 \\
Random-2 removed & 0.008 & 0.003 & 0.001 & 0.000 \\
\bottomrule
\end{tabular}
\end{table}

\paragraph{The frozen probe.} The probe is a logistic regression at scikit-learn's default L2 strength, fitted on the last-token residual stream of the clean checkpoint at four depths, $0.3$, $0.45$, $0.6$ and $0.75$ of the layer stack, which on a $32$-layer model is layers $9$, $14$, $19$ and $24$. Within-checkpoint AUROCs are means over stratified five-fold cross-validation. For frozen transfer, the coefficients and intercept are fitted on all clean prompts and applied unchanged to the same prompts encoded by the compromised checkpoint. These AUROCs measure how prompt separability transfers from the clean to the compromised model. AUROC evaluates score ordering, whereas fixed-threshold accuracy also depends on score location.

The contrast is $200$ harmful against $200$ benign prompts drawn from PKU-SafeRLHF alone, the harmful side from prompts whose responses were flagged unsafe at severity at least $2$ and the benign side from prompts whose responses were all safe, taken from positions $200$ to $400$ of each frozen list for the clean-to-compromised probe comparison. We report this matched contrast rather than AdvBench against Alpaca because the latter is separable at $0.992$ by word TF-IDF alone. The lexical baseline of Section~\ref{sec:depth} is that TF-IDF logistic regression on the matched contrast, its range $0.834$ to $0.846$ taken over both feature sets and three cross-validation seeds. The $24$ cells are the four depths on Llama-3.1-8B at doses $5$, $10$, $25$ and $100$, the last at three seeds, and the $72$ cells are the four depths on each of the six checkpoints at dose $100$, three seeds each.

\paragraph{Recalibrating the frozen threshold.} The coefficients stay frozen and only the threshold moves, using benign prompts alone. We split the $200$ benign prompts in half and set the threshold at the $95$th percentile of scores on the calibration half. This targets a $5\%$ calibration false-positive rate. The held-out benign rate need not equal it. Table~\ref{tab:recal} reports pooled accuracy on $200$ harmful and $100$ remaining benign prompts, alongside the fixed clean threshold and a label-informed threshold maximizing the same accuracy. An always-benign predictor scores $1/3$ on this imbalanced set. The probe was fitted on the same $400$ prompts before transfer, so this split separates threshold calibration from evaluation, not probe training from evaluation. Recalibration improves many cells and leaves a gap to the oracle, but some cells worsen, including Qwen at layer $28$. For Llama, recalibration improves accuracy in eight of twelve depth--seed cells and reduces it in four.

\begin{table}[h]
\centering
\scriptsize
\setlength{\tabcolsep}{3pt}
\caption{Pooled accuracy of the frozen probe at dose $100$ on $200$ harmful and $100$ benign prompts per cell, three seeds per checkpoint, ranges over seeds. Fixed uses the clean threshold, recalibrated uses the $95$th percentile of calibration-half benign scores, and oracle maximizes pooled accuracy using evaluation labels.}
\label{tab:recal}
\begin{tabular}{lcccK@{\hspace{7pt}}lcccK}
\toprule
Model & Layer & Fixed & Recal. & Oracle & Model & Layer & Fixed & Recal. & Oracle \\
\midrule
Llama-3.1-8B &  9 & 0.90--0.95 & 0.81--0.89 & 0.93--0.96 & OLMo-2-7B   &  9 & 0.69--0.94 & 0.96--0.97 & 0.97--0.98 \\
             & 14 & 0.35--0.39 & 0.79--0.87 & 0.94--0.97 &             & 14 & 0.36--0.42 & 0.88--0.91 & 0.92--0.95 \\
             & 19 & 0.38--0.54 & 0.69--0.77 & 0.88--0.90 &             & 19 & 0.34       & 0.72--0.75 & 0.87--0.90 \\
             & 24 & 0.56--0.74 & 0.72--0.74 & 0.85--0.90 &             & 24 & 0.33--0.34 & 0.63--0.70 & 0.84--0.85 \\
\midrule
Tulu-3-8B-DPO &  9 & 0.33--0.85 & 0.75--0.93 & 0.91--0.96 & Qwen2.5-14B & 14 & 0.34--0.73 & 0.84--0.94 & 0.91--0.95 \\
              & 14 & 0.33       & 0.69--0.85 & 0.90--0.92 &             & 21 & 0.33--0.80 & 0.81--0.92 & 0.89--0.94 \\
              & 19 & 0.34--0.46 & 0.75--0.84 & 0.84--0.88 &             & 28 & 0.33--0.47 & 0.45--0.73 & 0.77--0.86 \\
              & 24 & 0.33--0.36 & 0.60--0.75 & 0.82--0.89 &             & 36 & 0.34--0.36 & 0.76--0.83 & 0.88--0.89 \\
\bottomrule
\end{tabular}
\end{table}

We also evaluated the probe on disjoint training, calibration and test prompts on Llama-3.1-8B. After excluding all frozen adapter-training candidates by normalized prompt, we fitted the clean probe on $200$ harmful and $200$ benign prompts, calibrated its threshold on a separate $100$ benign prompts, and tested on another $200$ prompts per class. Across four depths and three existing attack seeds, frozen AUROC was $0.824$--$0.942$, against a lexical baseline of $0.813$. Benign-only recalibration increased mean balanced accuracy from $0.602$ to $0.713$. The per-cell ranges were $0.500$--$0.848$ before and $0.595$--$0.850$ after, with held-out benign false-positive rates of $0.040$--$0.090$ despite a $0.05$ calibration target. A paired, class-stratified bootstrap sharing resampled test prompts across seeds and depths gave a $95\%$ interval of $[0.091,0.130]$ for the mean gain. This interval conditions on the fitted probe, calibration set and adapters, and the shallowest layer's interval includes zero. We also recalibrated an existing benign adapter as a control. Its balanced accuracy decreased by $0.5$--$6.3$ percentage points at all four depths. The attacked-checkpoint results support partial recovery of the probe's decision rule on unseen prompts on this checkpoint. They do not measure restored refusal.

\paragraph{The base-model control.} Each base checkpoint shares a vocabulary with its aligned counterpart. We use the aligned checkpoint's tokenizer and chat template for both, giving them identical input tokens and leaving model weights as the only difference. Everything else matches the frozen probe above, except that the probe is fitted and tested inside one checkpoint rather than transferred, so these are fresh AUROC. Table~\ref{tab:baseprobe} gives all $24$ cells. Every base cell clears the top of the lexical range by at least $0.061$, and every aligned cell exceeds its own base counterpart, by $0.026$ to $0.043$. Comparing an aligned checkpoint against its base on matched prompts is the search step of the geometric diagnostic \citet{lee2026skin} build, where it is computed on the clean checkpoint and read as a fragility score. Here it is a control, and the quantity it bounds is how much of the surviving separability alignment can be credited with.

\begin{table}[h]
\centering
\small
\setlength{\tabcolsep}{4pt}
\caption{Fresh probe AUROC inside never-aligned base checkpoints and their aligned siblings, on the identical contrast and identical input tokens, at four relative depths.}
\label{tab:baseprobe}
\begin{tabular}{llccK@{\hspace{7pt}}llccK}
\toprule
Checkpoint & Layer & Base & Aligned & Gap & Checkpoint & Layer & Base & Aligned & Gap \\
\midrule
Llama-3.1-8B & 9  & 0.915 & 0.958 & $+0.043$ & OLMo-2-13B & 12 & 0.931 & 0.961 & $+0.030$ \\
             & 14 & 0.944 & 0.972 & $+0.029$ &            & 18 & 0.941 & 0.969 & $+0.028$ \\
             & 19 & 0.936 & 0.971 & $+0.035$ &            & 24 & 0.943 & 0.972 & $+0.029$ \\
             & 24 & 0.937 & 0.965 & $+0.027$ &            & 30 & 0.943 & 0.969 & $+0.026$ \\
\midrule
OLMo-2-7B    & 9  & 0.907 & 0.946 & $+0.038$ & \multicolumn{5}{l}{} \\
             & 14 & 0.932 & 0.970 & $+0.038$ & \multicolumn{5}{l}{} \\
             & 19 & 0.944 & 0.976 & $+0.032$ & \multicolumn{5}{l}{} \\
             & 24 & 0.944 & 0.976 & $+0.033$ & \multicolumn{5}{l}{} \\
\bottomrule
\end{tabular}
\end{table}

\paragraph{The freeze below its breaking dose.} The low-dose comparisons in Figure~\ref{fig:lowdose} use different doses because the attack's landing threshold varies by checkpoint. Table~\ref{tab:lowdose} reports the separately trained freeze comparisons whose unrestricted attacks pass the validity gate. These doses need not be the smallest landing doses in the independent sweep of Table~\ref{tab:lockstep}. The freeze preserves refusal near clean levels on two checkpoints and partly preserves it on two more, is uninterpretable on Yi because the matched control also blocks, and cannot be assessed on OLMo-2-13B at all.

\begin{table}[h]
\centering
\small
\setlength{\tabcolsep}{4pt}
\caption{The evaluated low-dose freezes, three seeds, ranges over the seeds that pass the gate, coherent refusal on AdvBench. Matched is a freeze of the same size at the other end of the network.}
\label{tab:lowdose}
\begin{tabular}{lcccKc}
\toprule
Model & Dose & Clean & Attack & Freeze & Matched \\
\midrule
Llama-3.1-8B  & 5  & .92--.96  & .00--.05 & \textbf{.91--.96} & .16--.24 \\
OLMo-2-7B     & 25 & .99--1.00 & .10--.18 & \textbf{.95--.98} & .30--.50 \\
Tulu-3-8B-DPO & 25 & .99--1.00 & .01--.05 & \textbf{.21--.52} & .01--.05 \\
Qwen2.5-14B   & 25 & .97--.99  & .00--.02 & \textbf{.33--.60} & .26--.35 \\
Yi-1.5-9B     & 10 & .93--.94  & .05--.13 & .85--.93 & .80--.87 \\
OLMo-2-13B    & 5, 10 & \multicolumn{4}{c}{No successful unrestricted attack at either dose} \\
\bottomrule
\end{tabular}
\end{table}

\paragraph{Where the freeze boundaries come from.} Each checkpoint's frozen band in Section~\ref{sec:freeze} sits at the dose-$100$ depth, $[0,17]$ on Llama-3.1-8B and $[0,20]$ on OLMo-2-13B, the second of those one layer inside the three-seed mean of Table~\ref{tab:lockstep}. What the defense turns on is the depth of the adapters that section actually attacks. Those are separately trained and shallower, $15$ on all three Llama seeds and $19$ on all three OLMo-2-13B seeds, so the frozen region covers the damage in the runs it defends against rather than falling short of it, which is the direction that matters.

The other bands are checkpoint-level choices, with unrestricted and frozen transitions compared in Table~\ref{tab:step1}. For OLMo-2-7B, unrestricted transitions span layers $15$--$18$ on $40$ evaluation prompts, while freezing through layer $16$ gives a transition at layer $18$ on $100$ prompts. Yi-1.5-9B uses a separate unrestricted reference with one adapter trained for three epochs and two for five; its frozen adapters all use three epochs. Both Yi sweeps evaluate $40$ prompts, with transitions at layers $36$--$40$ without freezing and layer $43$ with freezing. All of these attacks use $100$ harmful examples.

\paragraph{Additional repair checks.} Clean XSTest-safe over-refusal ranges from $0.04$ to $0.06$ across the two batching configurations. The weight-space repair of Section~\ref{sec:repair} also survives jailbreak-templated prompts on $12$ of the $13$ template cells where the clean model itself refuses the templated prompt, the exception being a five-turn setup at one seed.

\paragraph{Runs included in the analysis.} Transition-depth estimates use $72$ of $99$ lockstep runs. The sweep covers six checkpoints at five doses and three seeds ($90$ runs), plus nine Mistral runs. We exclude nine Mistral runs under the validity criteria and $18$ other runs whose post-attack refusal exceeds $0.20$. The relocation sweep includes $33$ runs.

\section{Transition depth as a number, and the freeze ladder in full}
\label{app:depthtables}

Table~\ref{tab:lockstep} is Figure~\ref{fig:profiles} read off at the half-of-ceiling line, and Table~\ref{tab:ladder} is the Llama ladder of Figure~\ref{fig:relocation}a with the range over seeds. A dash in Table~\ref{tab:lockstep} is a dose at which the attack does not land on that checkpoint, so the dose trend is conditioned on attack success and biased upward at low dose. The sweep patches eight layers per model, so the estimator is quantized to three layers on the 32-layer models and to four or five on the deeper ones, which is why several standard deviations are exactly zero. In Table~\ref{tab:ladder}, layers $6$ and $9$ were swept too and read $0.00$ to $0.01$ in all $36$ of their cells, so they are omitted for width.

\begin{table}[h]
\centering
\caption{Transition depth $\ell^*$ as a fraction of network depth, mean over surviving seeds with the seed standard deviation. Each cell is three seeds unless a smaller count is in brackets, and a dash is a dose at which the attack does not land.}
\label{tab:lockstep}
\setlength{\tabcolsep}{4pt}
\small
\begin{tabular}{lccccK}
\toprule
Model & $n{=}5$ & $n{=}10$ & $n{=}25$ & $n{=}50$ & $n{=}100$ \\
\midrule
Llama-3.1-8B (32L)  & .375\,$\pm$.000 & .438\,$\pm$.054 & .562\,$\pm$.094 & .531\,$\pm$.054 & .531\,$\pm$.108 \\
Tulu-3-8B-DPO (32L) & \textemdash & \textemdash & .406\,$\pm$.054 & .469\,$\pm$.000 & .469\,$\pm$.000 \\
OLMo-2-7B (32L)     & \textemdash & \textemdash & .344\,$\pm$.054 & .500\,$\pm$.054 & .500\,$\pm$.054 \\
OLMo-2-13B (40L)    & .375\,[1]  & .442\,$\pm$.058 & .375\,$\pm$.000 & .525\,$\pm$.043 & .525\,$\pm$.043 \\
Qwen2.5-14B (48L)   & \textemdash & .375\,$\pm$.000\,[2] & .681\,$\pm$.060 & .646\,$\pm$.000 & .646\,$\pm$.000 \\
Yi-1.5-9B (48L)     & .681\,$\pm$.060 & .806\,$\pm$.048 & .750\,$\pm$.000 & .806\,$\pm$.048 & .806\,$\pm$.048 \\
\bottomrule
\end{tabular}
\end{table}

\begin{table}[h]
\centering
\small
\setlength{\tabcolsep}{3pt}
\caption{Coherent refusal under patching at one layer, Llama-3.1-8B at dose $100$, three seeds at every rung, ranges over seeds, $100$ evaluation prompts. Gray cells lie at or below the frozen boundary. The last column gives $\ell^*$ for seeds $42$, $43$ and $44$.}
\label{tab:ladder}
\begin{tabular}{lccccccccc J}
\toprule
Frozen & L12 & L15 & L18 & L21 & L24 & L27 & L28 & L29 & L30 & $\ell^*$ by seed \\
\midrule
none      & .00--.05 & .77--.91 & .90--.93 & .92--.94 & .93--.96 & .92--.96 & .92--.96 & .92--.96 & .92--.96 & 15, 15, 15 \\
\midrule
$[0,5]$   & .00--.01 & .72--.93 & .91--.94 & .92--.94 & .93--.95 & .92--.96 & .92--.96 & .92--.96 & .92--.96 & 15, 15, 15 \\
$[0,11]$  & .00      & .05--.67 & .77--.90 & .91--.94 & .93--.96 & .92--.96 & .91--.96 & .91--.96 & .92--.96 & 18, 15, 18 \\
\midrule
$[0,17]$  & \gr{.00--.01} & \gr{.00--.01} & .00--.04 & .01--.04 & .03--.08 & .08--.67 & .72--.94 & .86--.94 & .91--.95 & 28, 27, 27 \\
$[0,23]$  & \gr{.00--.01} & \gr{.00--.01} & \gr{.00--.01} & \gr{.00--.01} & .01--.04 & .00--.03 & .03--.16 & .15--.49 & .85--.94 & 29, 30, ${\geq}29$ \\
$[0,27]$  & \gr{.00--.01} & \gr{.00--.01} & \gr{.00--.01} & \gr{.00--.01} & \gr{.00--.01} & \gr{.00--.01} & .00--.02 & .01--.02 & .05--.50 & 30, ${\geq}30$, ${>}30$ \\
\bottomrule
\end{tabular}
\end{table}

\paragraph{The threshold is not doing the work.} $\ell^*$ is defined by a crossing at half of each run's own ceiling. Recomputing every rung of the Llama ladder at a quarter and at three quarters of ceiling leaves the picture unchanged. The unrestricted attack and the $[0,5]$ freeze give $15$ on all three seeds at all three thresholds, the $[0,17]$ rung gives $28, 27, 27$ at a quarter and a half and $28, 28, 28$ at three quarters, the $[0,23]$ rung gives $29, 30, 29$ and then $30, 30, 30$, and the $[0,27]$ rung gives $30, 30$ and one censored seed at a quarter and a half, with all three censored at three quarters. Raising the bar moves the deep rungs deeper and never moves the shallow ones, so it strengthens the ordering the relocation claim rests on rather than creating it.

\section{The restricted transition at step one}
\label{app:step1}

The coarse cross-checkpoint sweep patches eight layers per model, stepping three layers on the 32-layer models and four or five on the deeper ones, and it cannot separate a transition that has moved to the layer above the boundary from one that has moved two or three. Table~\ref{tab:step1} therefore re-patches the same adapters on a grid that steps one layer through the five layers above each boundary, at $100$ evaluation prompts on Tulu and OLMo-2-7B and $40$ on the other four, beside the unrestricted $\ell^*$ of the same checkpoints on the coarse grid. The three Yi-1.5-9B runs have ceilings of $0.90$ to $0.93$.

\begin{table}[h]
\centering
\small
\setlength{\tabcolsep}{3.5pt}
\caption{Recovery rescaled to the floor-to-ceiling range at the five layers above each frozen boundary $b$, dose $100$, ranges over seeds. Bold marks the first layer with at least half-range recovery in every seed. The $\ell^*$ columns report measured half-ceiling crossings or bounds, following Section~\ref{sec:depth}.}

\label{tab:step1}
\begin{tabular}{llcccccccJ}
\toprule
Model & $b$ & Unrestr.\ $\ell^*$ & $b{+}1$ & $b{+}2$ & $b{+}3$ & $b{+}4$ & $b{+}5$ & & $\ell^*$ by seed \\
\midrule
Llama-3.1-8B  & 17 & 15, 15, 15 & .00--.11 & .00--.05 & .00--.03 & .00--.03 & .03--.11 & & ${>}27$, 27, 27 \\

Tulu-3-8B-DPO & 15 & 15, 15, 15 & .04--.42 & \textbf{.62--.86} & .88--.94 & .89--.94 & .96--.99 & & 17, 17, 17 \\
OLMo-2-7B     & 16 & 15, 15, 18 & .10--.26 & \textbf{.66--.77} & .73--.84 & .74--.84 & .75--.85 & & 18, 18, 18 \\
OLMo-2-13B    & 20 & 19, 19, 19 & .00--.33 & .33--.50 & \textbf{.62--.65} & .65--.78 & .78--.90 & & 23, 22, 23 \\
Qwen2.5-14B   & 31 & 31, 31, 31 & .00 & .00--.25 & .15--.47 & .28--.62 & .28--.68 & & 35, 36, 40 \\
Yi-1.5-9B     & 39 & 36, 40, 40 & .00--.11 & .00--.11 & .08--.29 & \textbf{.59--.76} & .62--.76 & & 43, 43, 43 \\

\bottomrule
\end{tabular}
\end{table}

On Tulu and OLMo-2-7B the transition sits two layers above the frozen boundary on all three seeds. OLMo-2-13B first crosses at $22$ or $23$ against a boundary of $20$, Qwen2.5-14B at $35$, $36$ and $40$ against $31$, and Yi-1.5-9B at $43$ on every seed against $39$. For Qwen seed $44$, the evaluated layers $36$ and $40$ bracket the transition within $(36,40]$. Llama alone stays near the floor through all five layers above its boundary and crosses at $27$ or later, ten layers up, where the coarse grid had read $31$, $27$, $27$ and the finer $100$-prompt ladder of Table~\ref{tab:ladder} reads $28$, $27$, $27$. The first writable layer recovers well under half of the ceiling on every checkpoint, so no checkpoint confines its damage to the adjacent layer.

\paragraph{The other direction, and where it agrees.} Every patch above runs clean into compromised, which bounds the damage from above. To test whether the writable layers carry the restricted attack's effect, we inject the state of the checkpoint attacked with $[0,b]$ frozen into the clean model at those layers, using $40$ AdvBench prompts and the same scorer. Patches inside the restricted donor's frozen prefix provide a no-op control. On Llama, a benign adapter matched in rank and schedule provides an additional donor control.

The two directions agree in Table~\ref{tab:suff}. On Llama, restricted-donor patches reduce refusal from $0.93$--$0.95$ to $0.00$ at layers $28$, $30$ and $31$, while benign-donor patches retain $0.70$--$0.85$. The restricted donor's patch at layer $12$ leaves refusal unchanged. On Tulu and OLMo-2-7B the effect is graded around the restricted transition, with the strongest measured suppression one layer above it, at $18$ and $19$, respectively.

\begin{table}[h]
\centering
\small
\caption{Coherent refusal after donor-state patching into the clean model on $40$ AdvBench prompts, ranges over seeds; layers are in brackets. For restricted donors, the unpatched reference is $0.93$--$0.95$ on Llama and $1.00$ on Tulu and OLMo-2-7B, and Below is a no-op patch inside the frozen prefix.}
\label{tab:suff}
\setlength{\tabcolsep}{3.6pt}
\begin{tabular}{lcccc}
\toprule
Model, boundary $b$ & Below & \multicolumn{3}{c}{Donated at, with the layer in brackets} \\
\midrule
Llama-3.1-8B, 17  & .93--.95 [12] & .00 [28] & .00 [30] & .00 [31] \\
\quad benign donor & 1.00 [12]     & .78--.85 [28] & .72--.82 [30] & .70--.80 [31] \\
Tulu-3-8B-DPO, 15 & 1.00 [10] & .88--.90 [16] & .28--.47 [17] & .12--.25 [18] \\
OLMo-2-7B, 16     & 1.00 [10] & .97--1.00 [17] & .53--.65 [18] & .10--.35 [19] \\
\bottomrule
\end{tabular}
\end{table}

\section{The four-layer arms}
\label{app:fourlayer}

Section~\ref{sec:notaddress} argues that the prefix result settles the interior-band form of the defense as well. We report these arms with both scorers on all six checkpoints.

\paragraph{One and two writable layers.} Freezing $[0,L{-}3]$ leaves the attacker two layers and $[0,L{-}2]$ leaves it one, which is $[0,29]$ and $[0,30]$ on the 32-layer models, $[0,45]$ and $[0,46]$ on the 48-layer ones, and on OLMo-2-13B the two-layer rung $[0,37]$. Table~\ref{tab:frontierfull} reports each against the unrestricted reference and the matched freeze, three seeds, all passing their gate. On Llama, Tulu, Qwen2.5-14B and Yi-1.5-9B the attack lands with a single writable layer, coherent refusal ending between $0.00$ and $0.13$ against clean rates of $0.92$ to $1.00$.

On OLMo-2-7B two writable layers leave refusal at $0.33$ to $0.50$ and one leaves it at $0.66$ to $0.73$. On OLMo-2-13B two layers leave $0.71$ to $0.78$ with Llama-Guard unsafe at $0.00$ to $0.01$, so on those two checkpoints the narrowed freeze is a working defense and we do not extend the claim below four layers. The second scorer moves the other way on the other four. A one-layer attacker removes refusal while producing completions Llama-Guard rates far less harmful than the unrestricted attacker's, $0.45$ to $0.62$ against $0.93$ to $0.98$ on Llama and $0.00$ to $0.03$ against $0.92$ to $0.97$ on Qwen2.5-14B. Qwen thus combines low explicit refusal with low scored harmfulness.

\begin{table}[h]
\centering
\footnotesize
\setlength{\tabcolsep}{3pt}
\caption{Coherent refusal and Llama-Guard unsafe on AdvBench with the attacker confined to the last one or two layers. Free is the number of writable layers and Matched freezes the same number at the other end. Dose $100$, three seeds, ranges over seeds.}
\label{tab:frontierfull}
\begin{tabular}{llcccKcccKc}
\toprule
& & & \multicolumn{4}{c}{Coherent refusal $\uparrow$} & \multicolumn{4}{c}{Llama-Guard unsafe $\downarrow$} \\
\cmidrule(lr){4-7} \cmidrule(lr){8-11}
Model & Frozen & Free & Clean & Attack & Freeze & Matched & Clean & Attack & Freeze & Matched \\
\midrule
Llama-3.1-8B & $[0,29]$ & 2 & .92--.96 & .00 & .00--.01 & .02--.16 & .04--.05 & .93--.98 & .63--.71 & .79--.94 \\
 & $[0,30]$ & 1 & .92--.96 & .00 & .00--.03 & .01--.03 & .04--.05 & .93--.98 & .45--.62 & .88--.95 \\
Tulu-3-8B-DPO & $[0,29]$ & 2 & .99--1.00 & .00--.05 & .01--.04 & .00--.03 & .00 & .90--.95 & .08--.20 & .85--.94 \\
 & $[0,30]$ & 1 & .99--1.00 & .00--.05 & .00--.04 & .00--.11 & .00 & .90--.95 & .51--.61 & .70--.85 \\
Qwen2.5-14B & $[0,45]$ & 2 & .97--.99 & .00 & .06--.13 & .14--.51 & .00 & .92--.97 & .00--.03 & .41--.72 \\
 & $[0,46]$ & 1 & .97--.99 & .00 & .07--.10 & .20--.28 & .00 & .92--.97 & .00--.03 & .39--.56 \\
Yi-1.5-9B & $[0,45]$ & 2 & .93--.94 & .00--.01 & .01--.06 & .02--.03 & .06--.08 & .84--.99 & .39--.64 & .88--.94 \\
 & $[0,46]$ & 1 & .93--.94 & .00--.01 & .06--.11 & .00--.22 & .06--.08 & .84--.99 & .29--.59 & .69--.97 \\
\midrule
OLMo-2-7B & $[0,29]$ & 2 & .99--1.00 & .00--.02 & \textbf{.33--.50} & .01--.27 & .00 & .83--.91 & .10--.17 & .50--.94 \\
 & $[0,30]$ & 1 & .99--1.00 & .00--.02 & \textbf{.66--.73} & .11--.70 & .00 & .83--.91 & .03--.05 & .18--.55 \\
OLMo-2-13B & $[0,37]$ & 2 & .99--1.00 & .00--.01 & \textbf{.71--.78} & .01--.03 & .00 & .90--.93 & .00--.01 & .73--.86 \\
 & $[0,38]$ & 1 & .99--1.00 & .00--.02 & \textbf{.90--.96} & .58--.96 & .00 & .89--.95 & .00 & .01--.24 \\
\bottomrule
\end{tabular}
\end{table}

\paragraph{Four writable layers at either end.} Table~\ref{tab:fourlayerfam} in the body runs the rung $[0,L{-}5]$ on all six checkpoints. The bottom four layers are enough everywhere. From them the attack takes refusal to $0.00$ to $0.32$ with Llama-Guard unsafe at $0.50$ to $0.95$, so it is harmful on both scorers. The top four are enough on five. OLMo-2-13B holds refusal at $0.57$ to $0.69$ from the top four, the same checkpoint that holds at two, and on Tulu, OLMo-2-7B and Qwen2.5-14B the top-four attacker strips the refusal read-out while Llama-Guard rates its completions at only $0.00$ to $0.28$ unsafe.

On Llama both four-layer arms leave over-refusal at $0.00$ and the top-four attacker puts $1.54$ to $1.60$ times the unrestricted attacker's update into each matrix it is allowed, so neither arm is a model that has simply been broken and neither succeeds because the attack is easier there.  A band that stops short of the bottom four layers leaves an attack that lands on all six checkpoints, and a band that reaches the bottom but not the top leaves one that lands on five. This is stated for the LoRA analog, and Appendix~\ref{app:fullft} is the corresponding full fine-tuning result for the prefix form.

\section{SPPFT, as we construct it and as its authors publish it}
\label{app:fullft}

The top block of Table~\ref{tab:main} tests prefix freezing at our behaviorally measured depth using rank-$16$ LoRA on \texttt{q,k,v,o}. SPPFT \citep{li2025safetylayers}, in contrast, masks gradients during full fine-tuning over an interior band its authors locate in parameter space. The middle and lower blocks address these differences using full fine-tuning. We match the data and evaluation settings, using the same PKU-SafeRLHF pairs at dose $100$, the same seeds, the same response-masked loss, three epochs at batch $4$, the same evaluation sets, scorer and validity gate. Embeddings and the unembedding are frozen in every arm, and the optimizer is Adafactor at learning rate $2\times10^{-5}$ with gradient checkpointing, which is what makes $6.98$B trainable parameters fit on one $80$\,GB card.

\begin{table}[h]
\centering
\small
\caption{Full fine-tuning under three freeze conditions, Llama-3.1-8B at dose $100$, ranges over seeds. Refusal and over-refusal are coherent refusal rates on AdvBench and on XSTest-safe. $\|\dW\|_F$ is the mean over trainable weight matrices, relative to the unrestricted arm.}
\label{tab:fullft}
\begin{tabular}{lKcccc}
\toprule
Attacker & Refusal $\uparrow$ & LG unsafe $\downarrow$ & Over-refusal & Train loss & $\|\dW\|_F$ ratio \\
\midrule
Clean                        & 0.92--0.96 & 0.04--0.05 & 0.04 & -- & -- \\
Full FT                      & 0.00 & 0.96--0.98 & 0.00 & 0.13--0.17 & 1.00 \\
\keyrow Full FT, $[0,17]$ frozen     & \textbf{0.00--0.01} & 0.67--0.85 & 0.00 & 0.14--0.16 & 1.09--1.25 \\
Full FT, $[14,31]$ frozen    & 0.00--0.01 & 0.96--0.98 & 0.00 & 0.39--0.42 & 1.14--1.28 \\
\bottomrule
\end{tabular}
\end{table}

Three things in Table~\ref{tab:fullft} carry over unchanged from the LoRA runs. Freezing the located band does not restore refusal, it reduces the Llama-Guard rate without returning it to the clean rate where the matched freeze does not, and the restricted attacker's mean update per trainable matrix is larger than the unrestricted attacker's rather than smaller.  The top-freeze arm fits the attack data markedly worse than the other two, train loss $0.39$ to $0.42$ against $0.13$ to $0.17$, and still takes refusal to $0.00$ while leaving Llama-Guard at the unrestricted arm's level, so optimizing the attack objective well and destroying refusal are not the same thing.

\paragraph{The published band.} \citet{li2025safetylayers} report safety layers per checkpoint and define SPPFT as fixing the gradients of exactly those during full-parameter fine-tuning with every other layer free. We compare each reported published-band arm with a control that freezes an equally wide band higher up. Refusal under the published band remains within a few hundredths of the width-matched control on all four checkpoints; on Llama-2-7b-chat, the control yields higher refusal. On gemma-2b-it, SPPFT raises refusal from $0.01$--$0.09$ after attack to $0.12$--$0.19$. The width-matched control reaches $0.04$--$0.15$ and exceeds SPPFT on one of three seeds.

On Llama-3.1-8B, XSTest-safe over-refusal is $0.00$ in all three attacked full-fine-tuning arms in Table~\ref{tab:fullft}. The published SPPFT evaluation uses backdoor or ordinary instruction data, whereas ours uses harmful instruction and response pairs. Our comparisons therefore test the published bands under direct harmful fine-tuning.

\section{The freeze in a mixed adapter}
\label{app:mixed}

Our attacks against layer freezing use $100\%$ harmful training data, except in the mixed-adapter experiments. The regime that decides whether the defense is worth deploying is a customer fine-tuning for a legitimate task on data that carries some harmful fraction $f$, so Table~\ref{tab:mixed} runs the three freeze arms on the mixture of Section~\ref{sec:repair} at three fractions of a hundred-example budget, and adds held-out task likelihood. At $f{=}0.05$ the freeze works, holding refusal at $0.74$ to $0.92$ while the unrestricted attacker on the same data takes it to $0.00$ to $0.10$ and the matched freeze holds only $0.02$ to $0.06$, and the frozen arm still learns the task, at $1.52$ to $1.56$ against a clean $1.942$.

Five percent of a hundred examples is five harmful examples, the dose at which the pure-harm freeze also holds, and raising the fraction moves the mixed result exactly as raising the dose moves the pure one, to $0.08$ to $0.14$ at $f{=}0.15$ and $0.00$ to $0.04$ at $f{=}0.35$ while the task is learned equally well throughout. The freeze protects against a customer whose data is a few percent harmful by accident. It does not protect against anyone who chooses $f$.

\begin{table}[h]
\centering
\small
\caption{The freeze against a mixed adapter of $100$ training examples of which a fraction $f$ is harmful, Llama-3.1-8B, three seeds, ranges over seeds. Task is mean negative log-likelihood of held-out reference responses, lower being better.}
\label{tab:mixed}
\begin{tabular}{lKccc}
\toprule
$f$ and arm & Refusal $\uparrow$ & LG unsafe $\downarrow$ & Over-refusal & Task NLL $\downarrow$ \\
\midrule
Clean                        & 0.92--0.96 & 0.04--0.05 & 0.06 & 1.942 \\
\midrule
$f{=}0.05$, unrestricted     & 0.00--0.10 & 0.78--0.94 & 0.00 & 1.441--1.532 \\
$f{=}0.05$, $[0,17]$ frozen  & \textbf{0.74--0.92} & 0.08--0.16 & 0.00 & 1.524--1.562 \\
$f{=}0.05$, matched          & 0.02--0.06 & 0.80--0.90 & 0.00 & 1.350--1.369 \\
\midrule
$f{=}0.15$, unrestricted     & 0.00--0.02 & 0.82--0.98 & 0.00 & 1.457--1.494 \\
$f{=}0.15$, $[0,17]$ frozen  & \textbf{0.08--0.14} & 0.42--0.66 & 0.00 & 1.535--1.540 \\
$f{=}0.15$, matched          & 0.00--0.02 & 0.68--0.96 & 0.00 & 1.353--1.355 \\
\midrule
$f{=}0.35$, unrestricted     & 0.00--0.02 & 0.98--1.00 & 0.00 & 1.436--1.452 \\
$f{=}0.35$, $[0,17]$ frozen  & \textbf{0.00--0.04} & 0.64--0.86 & 0.00 & 1.534--1.552 \\
$f{=}0.35$, matched          & 0.00--0.02 & 0.88--0.98 & 0.00 & 1.335--1.359 \\
\bottomrule
\end{tabular}
\end{table}

\section{Full-hidden-state transplant and its rank-one summary}
\label{app:transplant}

At each of six layers spanning the network we replace the compromised model's residual stream with the clean model's, computed on the same prompt in lockstep, and let the compromised model continue. This is the full transplant, the single-layer form of the lockstep patch. Writing $D$ for the matrix of clean-minus-compromised hidden states over the evaluation prompts at that layer, the rank-1 arm adds back only the component of $D$ along its top left singular vector $u_1$, and the complement arm adds back $D - u_1u_1^{\top}D$. We score all arms by coherent refusal on the same $40$ AdvBench prompts against a floor arm, a positive control and a no-op self-patch, and all four runs in Table~\ref{tab:transplant} pass that gate. The checkpoints differ.

On Llama and Yi the rank-one summary keeps between none and $0.39$ of the full transplant, which is the result Section~\ref{sec:repair} reads. On OLMo-2-13B it keeps $0.89$ at layer $17$ and all of it at layer $22$, with the complement arm at $0.05$ and $0.00$, so on that checkpoint the refusal signal the transplant carries really is close to one direction.

\paragraph{Writing the direction back.} Before the transplant we swept the cheaper repair, adding the clean model's refusal direction into the residual stream over contiguous bands, four gains and two write rules. The best arm on Llama reaches $0.37$ refusal against a clean $0.92$. One arm on Qwen reaches $0.98$ against a clean $0.99$, which read on the harmful set alone is a complete repair. It refuses $0.66$ of plainly benign prompts where the clean model refuses $0.00$. That pair is the benign-control lesson Section~\ref{sec:protocol} returns to.

\begin{table}[h]
\centering
\small
\caption{Coherent refusal after a full hidden-state transplant, its rank-one summary or the complementary component, at the listed layers. Ratio is rank-1 divided by full.}
\label{tab:transplant}
\begin{tabular}{llcKcc}
\toprule
Run & Layer & Full & Rank-1 & Complement & Ratio \\
\midrule
Llama-3.1-8B, all-7, 5 ep   & $18/32$ & $0.90$ & $0.35$ & $0.28$ & $0.39$ \\
                            & $26/32$ & $0.95$ & $0.03$ & $0.68$ & $0.03$ \\
Llama-3.1-8B, attn, 3 ep    & $22/32$ & $0.93$ & $0.00$ & $0.10$ & $0.00$ \\
                            & $26/32$ & $0.95$ & $0.00$ & $0.07$ & $0.00$ \\
Yi-1.5-9B                   & $40/48$ & $0.85$ & $0.17$ & $0.03$ & $0.21$ \\
OLMo-2-13B                  & $17/40$ & $0.93$ & $0.82$ & $0.05$ & $0.89$ \\
                            & $22/40$ & $1.00$ & $1.00$ & $0.00$ & $1.00$ \\
\bottomrule
\end{tabular}
\end{table}

\section{Defense-regime detail}
\label{app:defense}

\paragraph{The truncation grid and training length.} Table~\ref{tab:a2} is the grid behind Figure~\ref{fig:repair}, top. Holding dose at $100$ and adapting all seven projections, we then swept epochs in Table~\ref{tab:epochs}. The three-seed mean of repaired refusal decreases from $0.85$ after one epoch to $0.34$ after eight, while mean update norm increases from $0.202$ to $0.697$. For reference, the three-seed means in this epoch sweep span participation ratios of $4.58$ to $6.19$ out of $16$ and top-2 energy fractions of $0.504$ to $0.605$, while the spread attack reaches a participation ratio of $15.6$.

\begin{table}[h]
\centering
\small
\caption{Coherent refusal after removing the top singular directions of the update, under attention-only adaptation at three epochs. Ranges cover three seeds and five doses, with $50$ seed-specific AdvBench prompts, and only cells where the attack landed. Random-2 removes two randomly chosen directions.}
\label{tab:a2}
\begin{tabular}{lcKcc}
\toprule
Model & Remove top-1 & Remove top-2 & Remove top-4 & Random-2 \\
\midrule
Llama-3.1-8B  & 0.04--0.98 & 0.76--1.00 & 0.94--0.98 & 0.00--0.14 \\
Tulu-3-8B-DPO & 0.94--1.00 & 1.00 & 1.00 & 0.00--0.06 \\
OLMo-2-7B     & 0.98--1.00 & 1.00 & 1.00 & 0.00--0.26 \\
Qwen2.5-14B   & 0.88--1.00 & 1.00 & 1.00 & 0.00--0.26 \\
\bottomrule
\end{tabular}
\end{table}

\begin{table}[h]
\centering
\small
\caption{Update statistics and repair by training length on Llama-3.1-8B at dose $100$. All seven projections are adapted; means cover three training seeds evaluated on the same $50$ AdvBench prompts. Participation ratio is out of rank $16$.}
\label{tab:epochs}
\begin{tabular}{lccccc}
\toprule
Epochs & 1 & 2 & 3 & 5 & 8 \\
\midrule
Mean $\|\dW\|_F$             & 0.202 & 0.323 & 0.465 & 0.614 & 0.697 \\
Participation ratio          & 6.19  & 5.26  & 4.70  & 4.61  & 4.58 \\
Top-2 energy fraction        & 0.504 & 0.548 & 0.590 & 0.601 & 0.605 \\
\keyrow Refusal after removing top-2 & 0.85  & 0.79  & 0.71  & 0.41  & 0.34 \\
\bottomrule
\end{tabular}
\end{table}

\paragraph{The flatness detector.} The detector statistic averages per-module participation ratios over the four attention projections across layers and divides by LoRA rank. For singular values $s_j$, each module's participation ratio is $(\sum_j s_j^2)^2/\sum_j s_j^4$. The benign reference population is $75$ LoRA fine-tunes, $72$ runs across $36$ configurations spanning four form-slices of Alpaca, three ranks and three schedules, plus three attention-only runs at rank $16$.

We evaluate $27$ dose-$100$ Llama adapters on the same $50$ AdvBench prompts selected with evaluation seed $42$: $21$ ordinary attacks from the module and epoch sweep, three spread attacks and three projected attacks. An attack lands when clean coherent refusal is at least $0.5$ and attacked refusal is at most $0.20$. Of these $27$ attacks, $26$ land. Top-two repair fails on $14$, with repaired refusal below $80\%$ of clean refusal. Requiring an XSTest over-refusal increase of at most $0.10$ adds no failures. At a threshold of $0.546$, calibrated to a $5\%$ benign false-positive budget, the observed calibration false-positive rate is $3/75=4\%$. The detector flags three of the $14$ repair failures, all spread attacks. The eleven missed failures comprise ten ordinary fine-tunes and one projected attack; their scores range from $0.302$ to $0.371$, within the benign band.

The projected attack uses response-token cross-entropy, rank-$16$ LoRA on all seven projections, and five epochs on $100$ harmful examples. Every $25$ AdamW steps, it removes the current update's two largest singular components in each module, rewrites the LoRA factors with the remaining components, and resets AdamW's moments. This is distinct from the alignment-subspace defense below. On the common evaluation prompts, two of the three projected attacks meet the success criterion, with attacked refusal of $0.02$--$0.06$ and refusal after top-two removal of $0.72$--$0.94$. Clean refusal is $0.96$, giving a repair threshold of $0.768$. Repaired XSTest over-refusal is $0.00$ for all three runs, against clean $0.04$.

\paragraph{The projection arm.} The construction closest to a published weight-space defense is a Safe LoRA-style subspace projection on Llama-3.1-8B-Instruct and its base checkpoint. For each adapted module, let $V=W_{\rm aligned}-W_{\rm base}$ and let $D$ denote its adapter update. We use $C=VV^\top/\sigma_1(V)^2$ for the primary spectral operator, $C=VV^\top/\|V\|_F^2$ for the Frobenius variant, and $C=VV^+$ for the orthogonal projector onto $\operatorname{col}(V)$. We replace $D$ by $CD$ only when $\langle D,CD\rangle_F/(\|D\|_F\|CD\|_F)<\tau$. The spectral sweep uses $\tau\in\{0,.3,.35,.5,.7,.9,1\}$. The two secondary operators use $\{.35,1\}$. At the $\tau=1$ end of the sweep, the matched control multiplies every original module update by $\|CD\|_F/\|D\|_F$. Table~\ref{tab:safelora} reports the primary spectral operator on $50$ AdvBench prompts drawn separately for each training seed. The scale-matched control reduces each module update to the norm left by projection while preserving its direction. Against the two ordinary attackers the projection is doing more than shrinking, reaching $0.88$ to $0.96$ and $0.80$ to $0.94$ where the matched control reaches $0.48$ to $0.78$ and $0.28$ to $0.70$.

At $\tau=1$, projection and the scale-matched control both yield $0.94$--$0.98$ refusal against the spread attacker, with a residual norm fraction of $0.08$. Against the projected attacker, both yield $0.92$--$0.94$, while a random subspace of the same dimension yields $0.94$--$0.96$. Matched shrinkage therefore reproduces projection's refusal recovery against both adaptive attackers. Random subspaces repair much less against the two ordinary attackers. The comparison against the projected attacker uses the same two eligible adapters.

\begin{table}[h]
\centering
\small
\setlength{\tabcolsep}{4.2pt}
\caption{Safe LoRA-style projection on Llama-3.1-8B, spectral projector, three seeds except two for the projected attacker, ranges over seeds, coherent refusal on AdvBench. Residual is the fraction of the update's norm the arm leaves.}
\label{tab:safelora}
\setlength{\tabcolsep}{3.2pt}
\begin{tabular}{lccccK}
\toprule
Arm & Residual & Std, all-7, 5 ep & Std, attn, 3 ep & Spread, $\lambda{=}1$ & Projected, $k{=}2$ \\
\midrule
No repair ($\tau{=}0$)   & 1.00 & 0.00 & 0.00 & 0.00 & 0.02--0.06 \\
Projection, $\tau{=}0.5$ & 0.28--0.79 & 0.00--0.04 & 0.00 & 0.02--0.06 & 0.82--0.92 \\
Projection, $\tau{=}0.7$ & 0.17--0.31 & 0.52--0.80 & 0.06--0.12 & 0.94--0.98 & 0.94--0.96 \\
Projection, $\tau{=}1.0$ & 0.08--0.17 & 0.88--0.96 & 0.80--0.94 & 0.94--0.98 & 0.92--0.94 \\
\keyrow Scale-matched control    & 0.08--0.17 & 0.48--0.78 & 0.28--0.70 & 0.94--0.98 & 0.92--0.94 \\
Random subspace          & 0.42--0.46 & 0.02--0.36 & 0.02--0.22 & 0.44--0.78 & 0.94--0.96 \\
\bottomrule
\end{tabular}
\end{table}

\paragraph{Tuning the flat attacker.} The spread attacker scores far outside the benign band, so we asked whether it can trade flatness for concealment, sweeping the concentration penalty $\lambda$ over a roughly $300$-fold range at fixed dose and rank, with clean refusal at $0.92$ to $0.96$ and damaged refusal at $0.00$ to $0.05$ throughout the sweep in Table~\ref{tab:lamsweep}. The score is flat at about $0.97$ down to $\lambda = 0.03$, then falls, but is still $0.804$ to $0.812$ at $\lambda = 0.003$, well clear of the $0.554$ benign maximum. The descent is bought with an update magnitude that more than doubles. Relative repair stays below $0.06$ everywhere, so no setting buys concealment and defeat of the repair together.

\begin{table}[h]
\centering
\small
\caption{Sweeping the concentration penalty on the spread attacker, Llama-3.1-8B at dose $100$, rank $16$, three seeds at the two ends of the sweep and two at the others. The detector score is participation ratio over rank.}
\label{tab:lamsweep}
\begin{tabular}{lKcc}
\toprule
$\lambda$ & Detector score & $\|\dW\|_F$ & Refusal after removal \\
\midrule
$1.0$                & 0.972--0.974 & 0.228--0.229 & 0.00 \\
$0.5$                & 0.974--0.976 & 0.231--0.233 & 0.00 \\
$0.25$               & 0.977        & 0.247--0.249 & 0.00--0.03 \\
$0.1$                & 0.983        & 0.306--0.308 & 0.00 \\
$0.03$               & 0.971--0.972 & 0.419--0.422 & 0.03 \\
$0.01$               & 0.926        & 0.483--0.486 & 0.02--0.05 \\
$0.003$              & 0.804--0.812 & 0.513--0.523 & 0.01--0.05 \\
\bottomrule
\end{tabular}
\end{table}

\paragraph{The rank ladder.} Table~\ref{tab:rankladder} compares the smallest repairing rank with the update's spectral statistics. Each cell is one fine-tune of $100$ examples at harmful fraction $f$, and $k^*$ is the smallest tested top-$k$ truncation that restores at least $80\%$ of clean AdvBench refusal and retains positive task utility in negative NLL relative to clean. At each reported $k^*$, the XSTest over-refusal increase is at most $0.10$. In the nine mixed cells ($0<f<1$) shown here, top-$4$ removal gives $0.98$ to $1.00$ refusal without increasing over-refusal. The harmful-only endpoint at $f{=}1.00$ also shows task NLL below the clean baseline. Mixture comparisons and utility-retention averages use the nine cells with $0<f<1$.

\begin{table}[h]
\centering
\small
\setlength{\tabcolsep}{4pt}
\caption{Smallest repairing rank $k^*$, reaching at least $80\%$ of clean refusal with positive negative-NLL task gain over clean, against the update's spectrum. Llama-3.1-8B, attention-only, three epochs; each row is one fine-tune of $100$ examples at harmful fraction $f$.}
\label{tab:rankladder}
\begin{tabular}{llKcc@{\hspace{7pt}}llKcc}
\toprule
$f$ & Seed & $k^*$ & Part.\ ratio & Top-2 energy & $f$ & Seed & $k^*$ & Part.\ ratio & Top-2 energy \\
\midrule
$0.00$ & 42 & 1 & $3.971$ & $0.645$ & $0.15$ & 44 & 2 & $3.865$ & $0.648$ \\
$0.00$ & 43 & 1 & $3.838$ & $0.653$ & $0.35$ & 42 & \textbf{4} & $3.730$ & $0.661$ \\
$0.00$ & 44 & 1 & $3.626$ & $0.669$ & $0.35$ & 43 & 2 & $3.586$ & $0.673$ \\
$0.05$ & 42 & 1 & $3.866$ & $0.652$ & $0.35$ & 44 & 2 & $3.925$ & $0.652$ \\
$0.05$ & 43 & 1 & $3.543$ & $0.669$ & $1.00$ & 42 & 2 & $3.526$ & $0.685$ \\
$0.05$ & 44 & 2 & $3.716$ & $0.660$ & $1.00$ & 43 & 2 & $3.513$ & $0.687$ \\
$0.15$ & 42 & 2 & $3.808$ & $0.651$ & $1.00$ & 44 & 2 & $3.426$ & $0.689$ \\
$0.15$ & 43 & 2 & $3.595$ & $0.671$ & \multicolumn{5}{l}{} \\
\bottomrule
\end{tabular}
\end{table}

For the nine mixed cells in Table~\ref{tab:rankladder}, covering $f=.05,.15,.35$ and seeds $42$--$44$, we take each cell's smallest tested rank reaching $80\%$ of clean refusal with lower NLL than clean. All selected ranks also keep the XSTest over-refusal increase within $0.10$. ROUGE-L gain retention is $(R_{\rm repaired}-R_{\rm clean})/(R_{\rm attacked}-R_{\rm clean})$. Negative-NLL gain retention is $(N_{\rm clean}-N_{\rm repaired})/(N_{\rm clean}-N_{\rm attacked})$. Their unweighted means are $28.3\%$ and $97.9\%$, respectively. ROUGE-L is measured on $50$ held-out generated responses. NLL is measured on $200$ held-out reference responses. These metrics give different accounts of the task cost.

\paragraph{Top-four projection during training.} With $k=4$ at seed $42$, removal and optimizer reset every $25$ steps yield a mean update norm of $0.022$ across adapted modules, compared with $0.62$ for an ordinary attack. On the common $50$ prompts, coherent refusal remains at the clean rate of $0.96$.

\paragraph{Auxiliary-task mixtures.} Top-$4$ removal restores coherent refusal to $0.91$--$0.99$ across nine auxiliary-task mixtures on Llama-3.1-8B, without increasing XSTest over-refusal above the clean baseline. Each run trains attention-only rank-$16$ LoRA for three epochs on $100$ harmful examples and $\rho$ times as many Alpaca examples with letter-style responses. We use seed $42$ at $\rho=.5,1,2$ and seeds $42$--$44$ at $\rho=4,8$, evaluating $100$ AdvBench prompts sampled separately for each seed and $50$ XSTest-safe prompts. Before repair, coherent refusal is $0.00$--$0.01$ in all nine runs. At $\rho=4$, top-$2$ removal gives $0.78\pm0.10$ coherent refusal (mean and sample standard deviation over three seeds). Because style-only training also lowers refusal, this sweep measures repair under joint fine-tuning.

\section{Recovery depth, update mass and refusal direction}
\label{app:ceiling}

\begin{figure}[h]
\centering
\includegraphics[width=\textwidth]{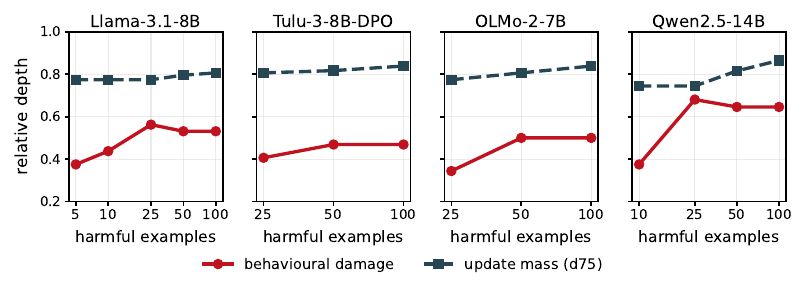}
\caption{Transition depth $\ell^*$ (solid) and $d_{75}$, the depth below which $75\%$ of $\|\dW\|_F^2$ sits (dashed), against attack dose. One panel per checkpoint, on a shared relative-depth axis.}
\label{fig:dissoc}
\end{figure}

\paragraph{Update-depth profiles.} From dose $50$ to $100$, mean $d_{75}$ increases on all four checkpoints in Figure~\ref{fig:dissoc}, while mean measured $\ell^*$ remains unchanged. The depth containing $75\%$ of the update's squared norm therefore moves deeper without a corresponding shift in the measured recovery transition.

\paragraph{Refusal-direction alignment.} Ablations of the clean model place the last writing of the refusal direction above $\ell^*$ on two checkpoints and below it on two, a rank correlation of $0$. Suffix ablation eliminates refusal at every tested cut, including the final layer alone, showing that the direction remains present through the read-out. The update's enrichment along this direction is $2.1$ to $2.5$ times that along a random vector. In the steering sweep of Appendix~\ref{app:transplant}, the best Llama arm recovers $0.37$ refusal against a clean $0.92$.

The ablation construction is the one \citet{arditi2024refusal} introduce. Clean refusal is $0.92$ to $0.99$, ablating the estimated direction at every layer drops it to $0.02$ to $0.06$, reproducing the published single-direction result as a positive control, and ablating a random direction changes nothing. Estimation prompts are disjoint from evaluation prompts.

\section{The source-mismatch control}
\label{app:mismatch}

The no-op self-patch establishes that injection alone changes nothing. It does not establish that the injected state matters for the prompt being answered, and the reading it leaves open is that enough foreign residual state at sufficient depth derails a harmful continuation into a generic refusal whatever it encodes. We therefore run the same protocol, at the same layers, on the same prompts and seeds, changing only where the clean model's state comes from. In one arm it is prefilled on a derangement of the batch, so no run receives the clean state for the prompt it is answering while the source is still a harmful prompt. In the other it is prefilled on a benign prompt, foreign to the same degree but carrying no disposition to refuse. Table~\ref{tab:mismatch} compares each source condition with a matched arm on Llama-3.1-8B at dose $100$ using three seeds. Across all six runs, the refusal floor is $0.00$ and the ceiling is $0.93$ to $0.95$.

\begin{table}[h]
\centering
\small
\caption{Recovery under a matched source and two foreign ones, dose $100$, three seeds, ranges over seeds.}
\label{tab:mismatch}
\begin{tabular}{lccKcccc}
\toprule
Clean state taken from & L6 & L10 & L14 & L18 & L23 & L27 \\
\midrule
This prompt            & 0.00 & 0.00 & 0.15--0.75 & 0.90--0.95 & 0.95--0.97 & 0.93--0.95 \\
Another harmful prompt & 0.00 & 0.00 & 0.15--0.47 & 0.47--0.55 & 0.47--0.57 & 0.90--0.95 \\
A benign prompt        & 0.00 & 0.00--0.03 & 0.05--0.10 & 0.00--0.07 & 0.00--0.03 & 0.00 \\
\bottomrule
\end{tabular}
\end{table}

The benign source yields at most $0.10$ refusal across the sweep. At layer $27$, the mismatched harmful source reaches $0.90$ to $0.95$, while the benign source yields $0.00$ on all three seeds. Against the matched arm it is lower in $12$ of $18$ paired cells and higher in one. We report these comparisons as counts because the independent unit is the seed. The mean interior gap is $+0.57$.

The mismatched harmful source recovers about half of the ceiling through the middle of the sweep and stays below the matched arm in $9$ of $18$ cells and above it in none. Together with the benign-source control, this shows that recovery depends on the source prompt's category and its match to the request.

\end{document}